%% file: egpaper_final.tex
\documentclass[10pt,twocolumn,letterpaper]{article}

\usepackage{iccv}
\usepackage{times}
\usepackage{enumitem}
\usepackage{epsfig}
\usepackage{subfig}
\usepackage{graphicx}
\usepackage{amsmath}
\usepackage{amssymb}
\usepackage{colortbl}
\usepackage{hhline}
\usepackage{makecell}
\usepackage{comment}
\usepackage{authblk}
\usepackage{multirow}
\usepackage{multicol}

\usepackage[breaklinks=true,bookmarks=false]{hyperref}

\iccvfinalcopy 

\def\iccvPaperID{3068} 

\ificcvfinal\fi
\begin{document}

\title{SpaceNet MVOI: a Multi-View Overhead Imagery Dataset}
\author[1]{Nicholas Weir}
\author[2]{David Lindenbaum}
\author[3]{Alexei Bastidas}
\author[1]{Adam Van Etten}
\author[3]{Sean McPherson}
\author[1]{Jacob Shermeyer}
\author[3]{Varun Kumar}
\author[3]{Hanlin Tang}
\affil[1]{In-Q-Tel CosmiQ Works, [nweir, avanetten, jshermeyer]@iqt.org}
\affil[2]{Accenture Federal Services, david.lindenbaum@accenturefederal.com}
\affil[3]{Intel AI Lab, [alexei.a.bastidas, sean.mcpherson, varun.v.kumar, hanlin.tang]@intel.com}

\maketitle
\thispagestyle{empty}

\begin{abstract}
\vspace{-10pt}
Detection and segmentation of objects in overheard imagery is a challenging task. The variable density, random orientation, small size, and instance-to-instance heterogeneity of objects in overhead imagery calls for approaches distinct from existing models designed for natural scene datasets. Though new overhead imagery datasets are being developed, they almost universally comprise a single view taken from directly overhead (``at nadir"), failing to address a critical variable: look angle. By contrast, views vary in real-world overhead imagery, particularly in dynamic scenarios such as natural disasters where first looks are often over $40^\circ$ off-nadir. This represents an important challenge to computer vision methods, as changing view angle adds distortions, alters resolution, and changes lighting. At present, the impact of these perturbations for algorithmic detection and segmentation of objects is untested. To address this problem, we present an open source Multi-View Overhead Imagery dataset, termed SpaceNet MVOI, with 27 unique looks from a broad range of viewing angles ($-32.5^\circ$ to $54.0^\circ$). Each of these images cover the same 665 $km^2$ geographic extent and are annotated with 126,747 building footprint labels, enabling direct assessment of the impact of viewpoint perturbation on model performance. We benchmark multiple leading segmentation and object detection models on: (1) building detection, (2) generalization to unseen viewing angles and resolutions, and (3) sensitivity of building footprint extraction to changes in resolution. We find that state of the art segmentation and object detection models struggle to identify buildings in off-nadir imagery and generalize poorly to unseen views, presenting an important benchmark to explore the broadly relevant challenge of detecting small, heterogeneous target objects in visually dynamic contexts.
\end{abstract}


\section{Introduction}

Recent years have seen increasing use of convolutional neural networks to analyze overhead imagery collected by aerial vehicles or space-based sensors, for applications ranging from agriculture \cite{Lacar2001} to surveillance \cite{Yuen2010,Uzkent:2017jv} to land type classification \cite{Chen:2015jd}. Segmentation and object detection of overhead imagery data requires identifying small, visually heterogeneous objects (e.g. cars and buildings) with varying orientation and density in images, a task ill-addressed by existing models developed for identification of comparatively larger and lower-abundance objects in natural scene images. The density and visual appearance of target objects change dramatically as look angle, geographic location, time of day, and seasonality vary, further complicating the problem. Addressing these challenges will provide broadly useful insights for the computer vision community as a whole: for example, how to build segmentation models to identify low-information objects in dense contexts.

\begin{figure*}[t]
\begin{center}
\vspace{-10pt}
\setlength{\tabcolsep}{0.15em}
\begin{tabular}{cccccc}

\vspace{-5pt}

& &
\textbf{Urban} &
\textbf{Industrial} &
\textbf{Dense Residential} &
\textbf{Sparse Residential} \\

\vspace{-5pt}

&
\raisebox{0.63in}{\rotatebox[origin=c]{90}{7 (NADIR)}} &
\subfloat{\includegraphics[width=0.225\linewidth]{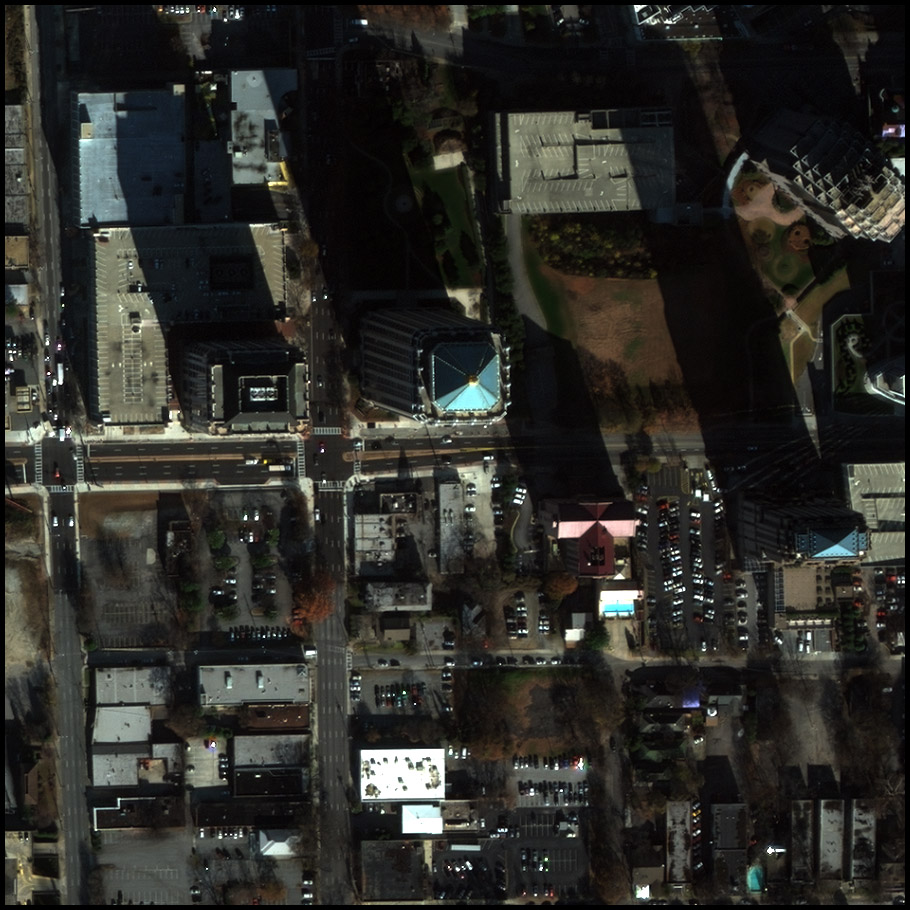}} &
\subfloat{\includegraphics[width=0.225\linewidth]{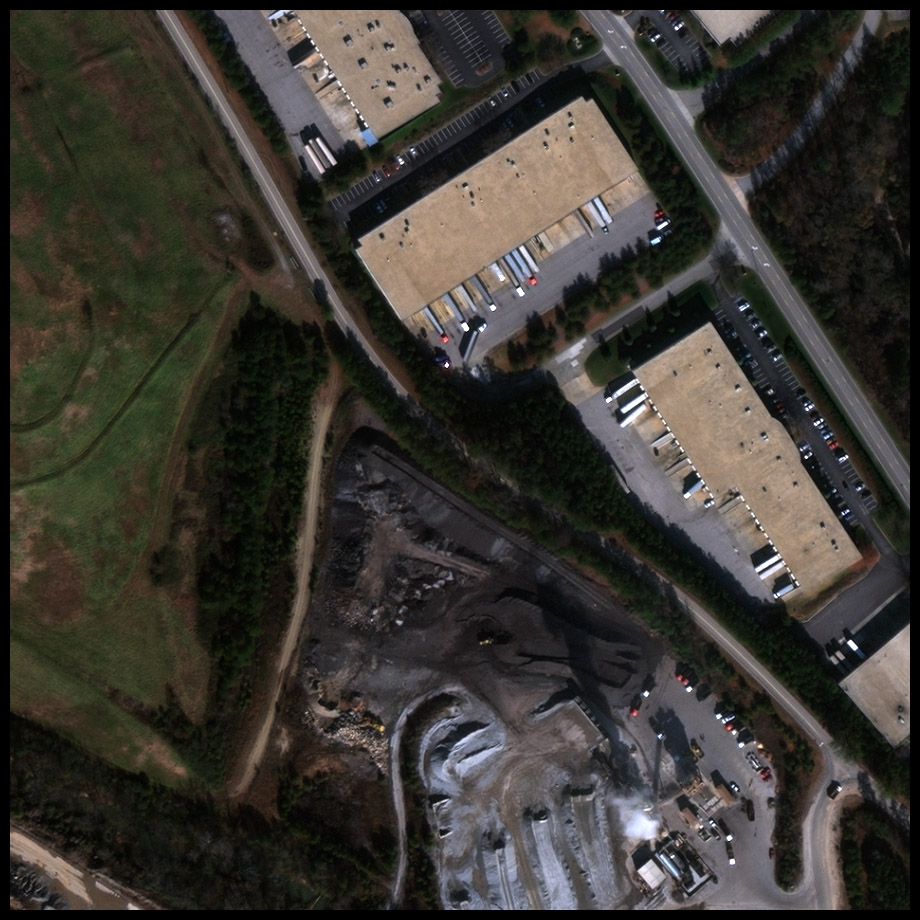}} &
\subfloat{\includegraphics[width=0.225\linewidth]{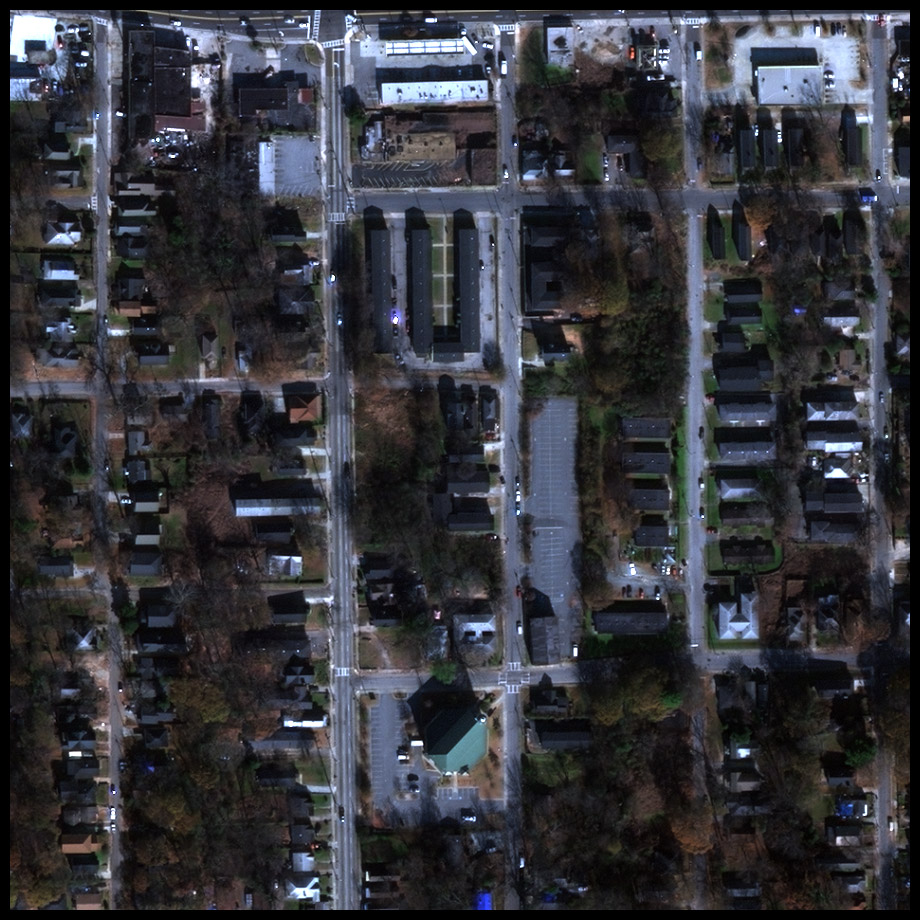}} &
\subfloat{\includegraphics[width=0.225\linewidth]{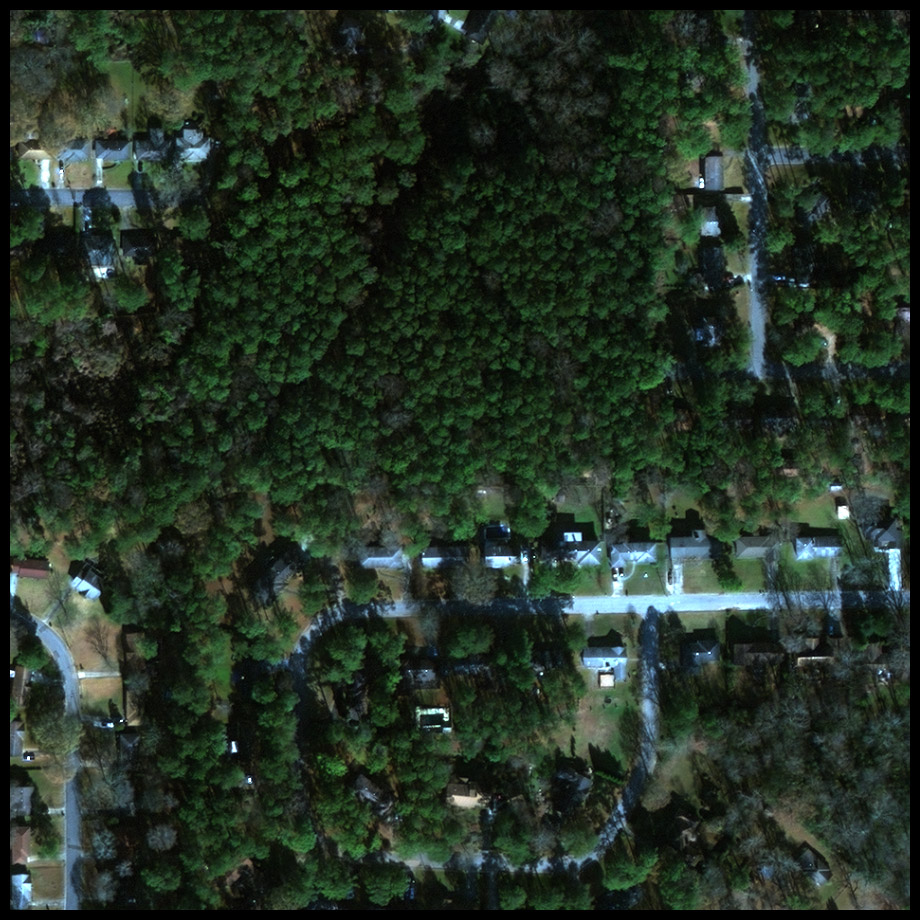}} \\

\vspace{-10pt}

\raisebox{0.7in}{\rotatebox[origin=c]{90}{\bf{LOOK ANGLE (BIN)}}} &
\raisebox{0.63in}{\rotatebox[origin=c]{90}{-32 (OFF)}} &
\subfloat{\includegraphics[width=0.225\linewidth]{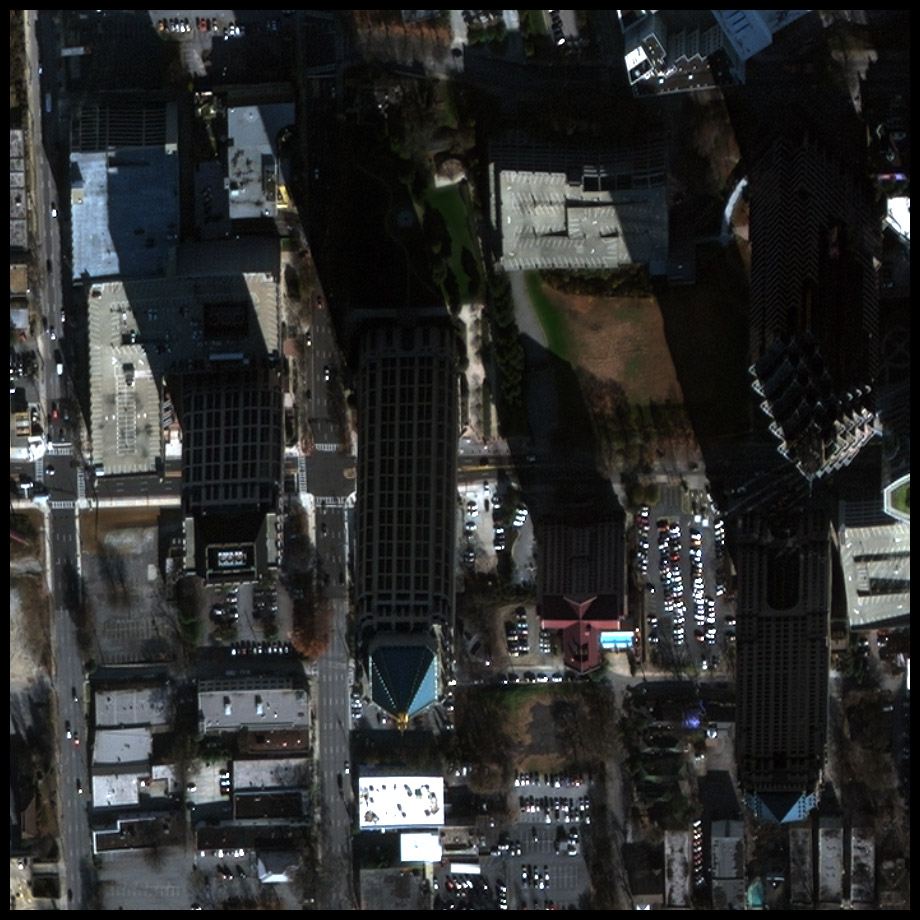}} &
\subfloat{\includegraphics[width=0.225\linewidth]{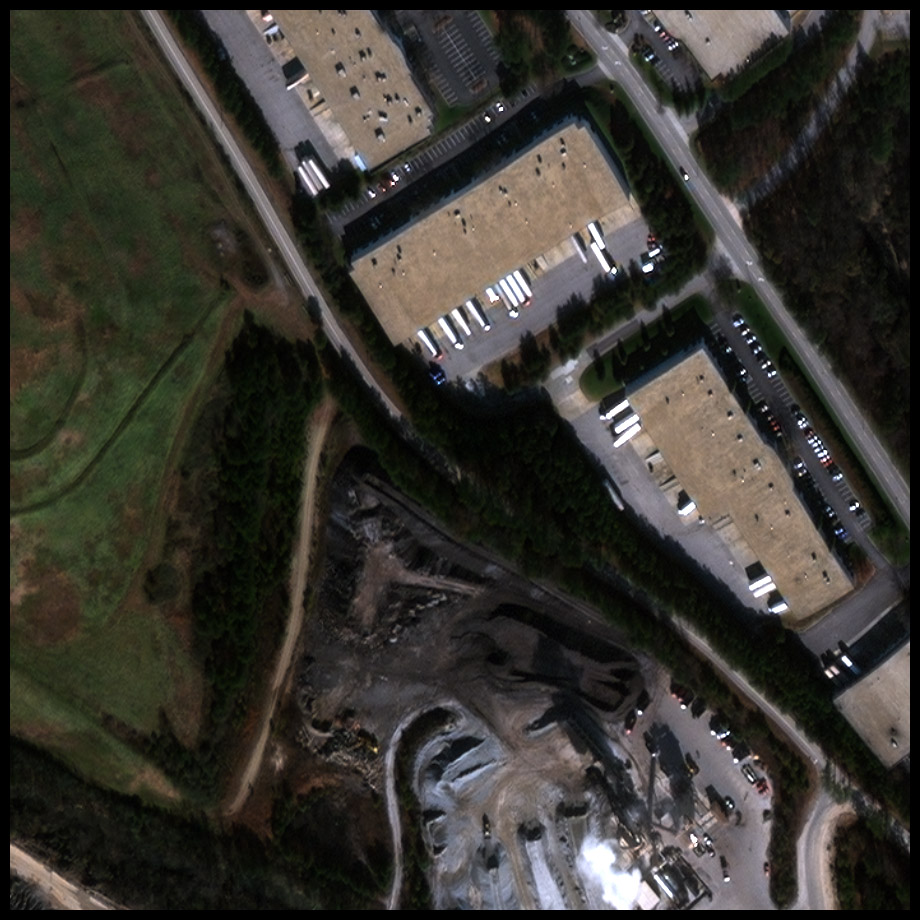}} &
\subfloat{\includegraphics[width=0.225\linewidth]{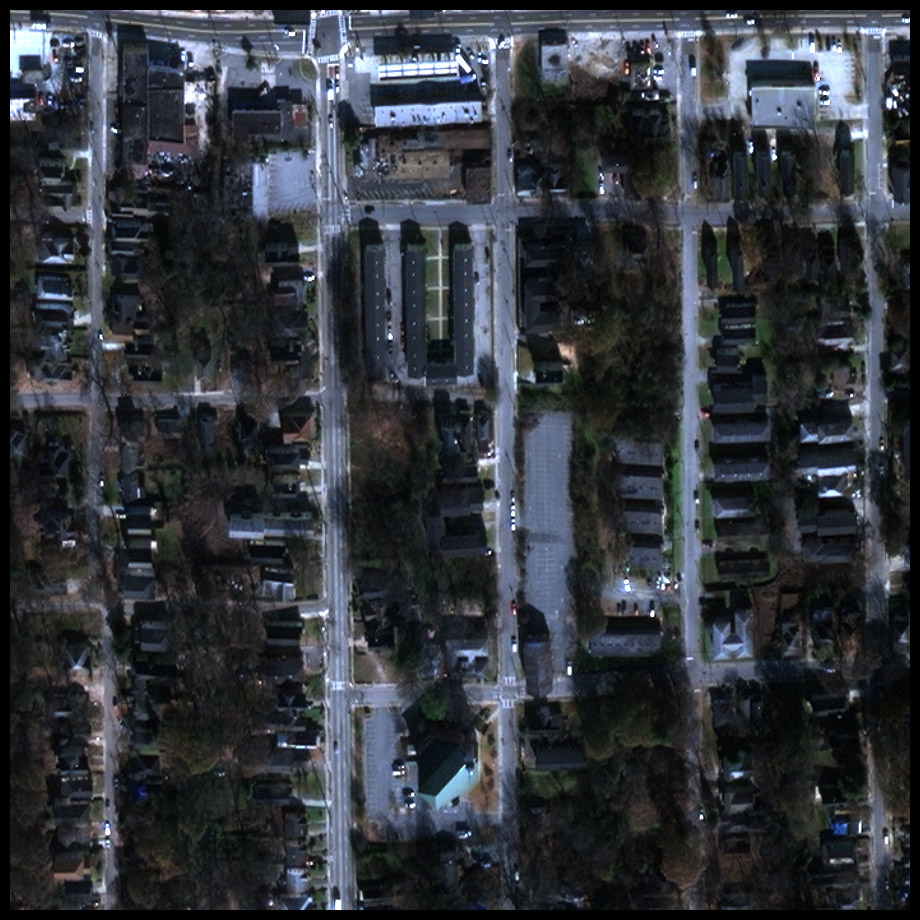}} &
\subfloat{\includegraphics[width=0.225\linewidth]{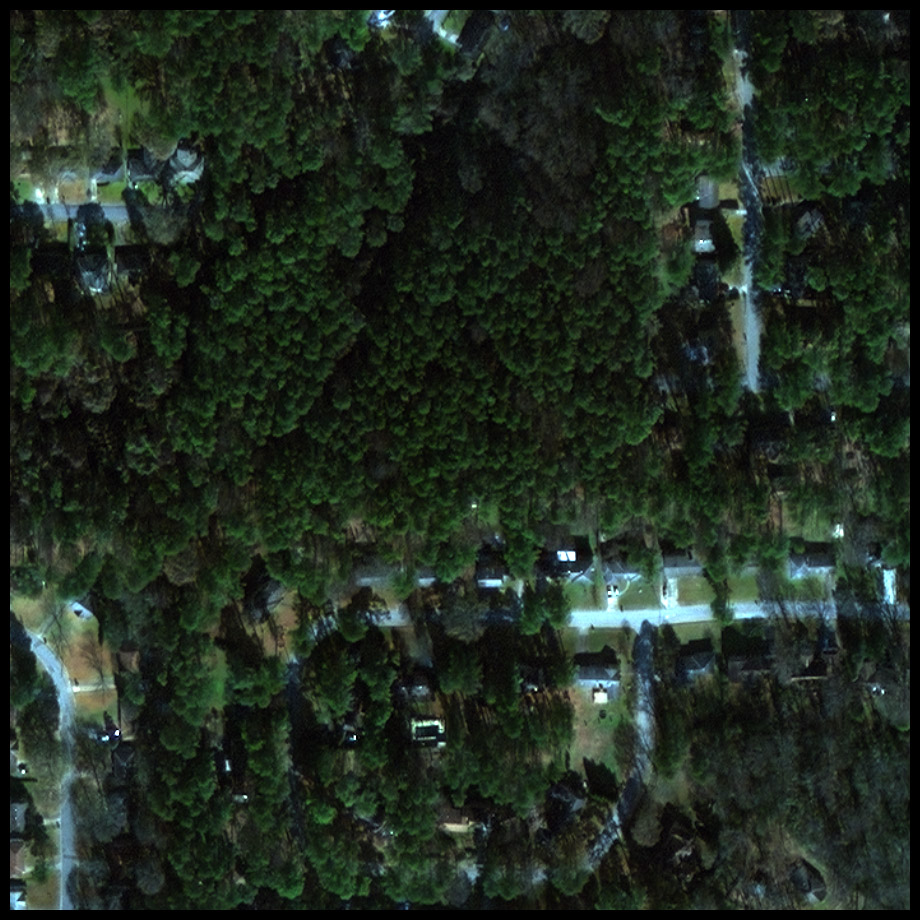}} \\

&
\raisebox{0.63in}{\rotatebox[origin=c]{90}{52 (VOFF)}} &
\subfloat{\includegraphics[width=0.225\linewidth]{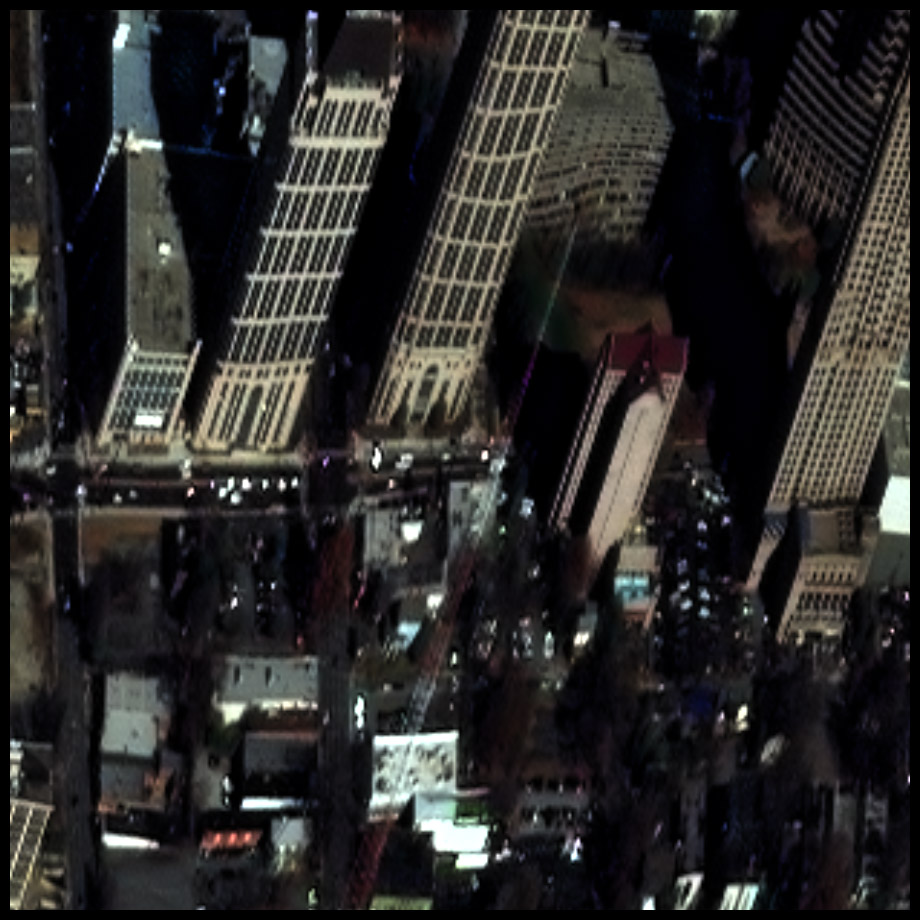}} &
\subfloat{\includegraphics[width=0.225\linewidth]{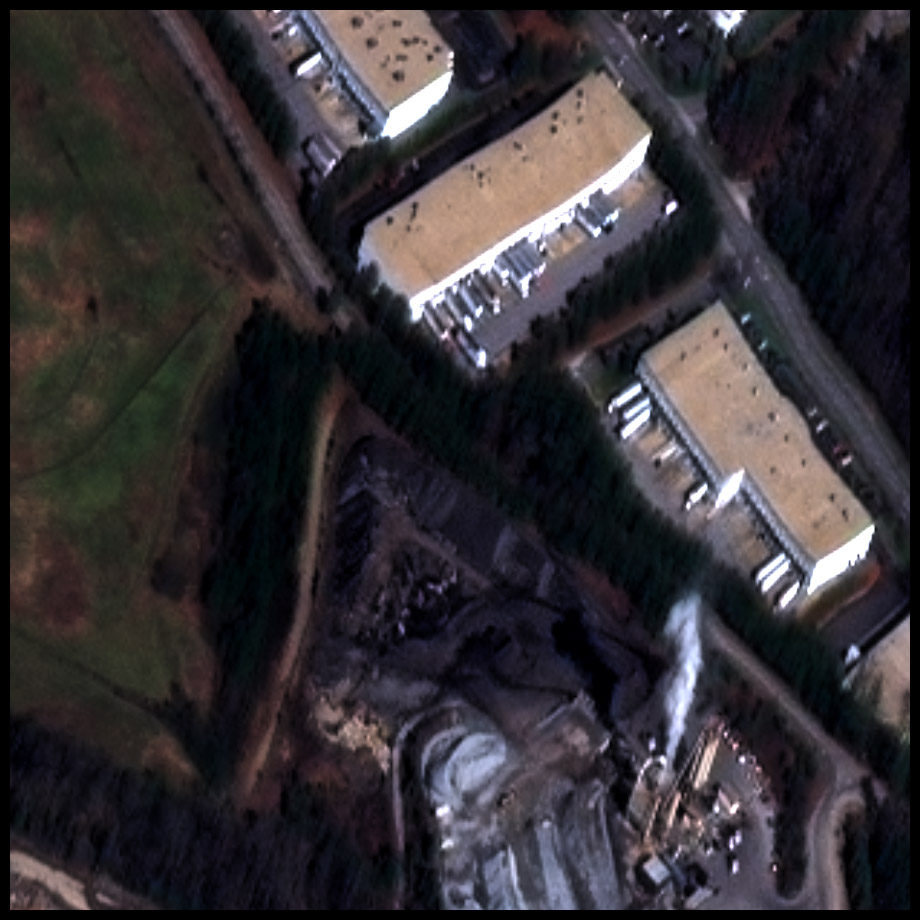}} &
\subfloat{\includegraphics[width=0.225\linewidth]{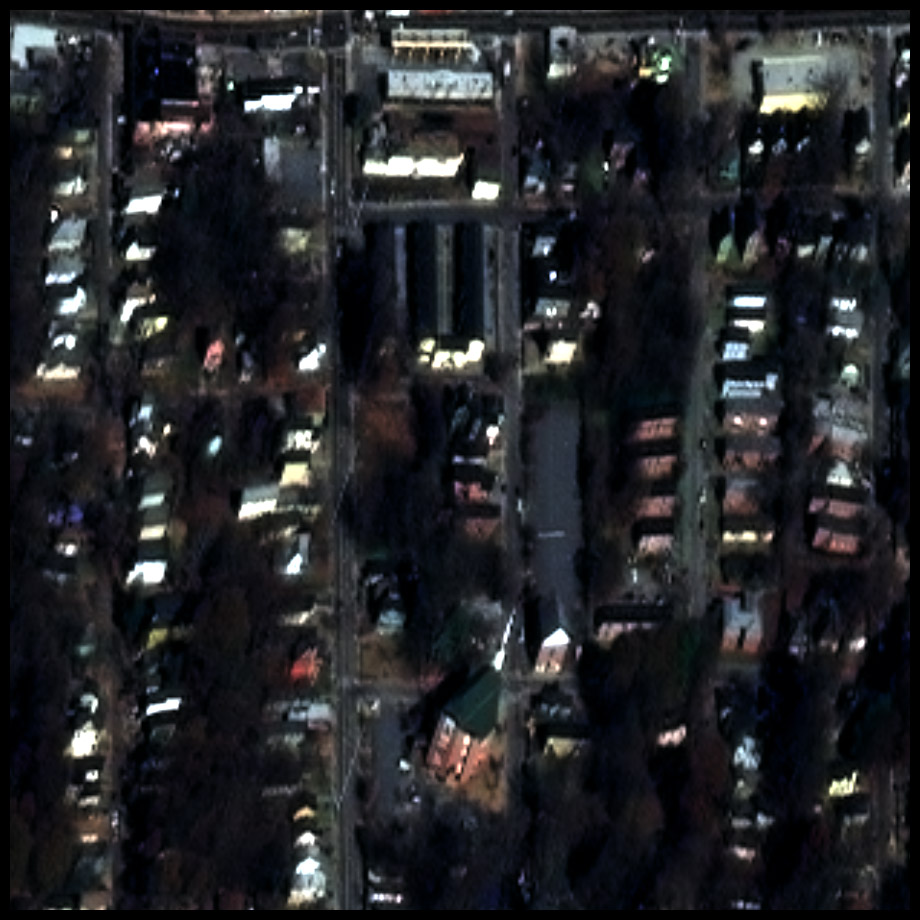}} &
\subfloat{\includegraphics[width=0.225\linewidth]{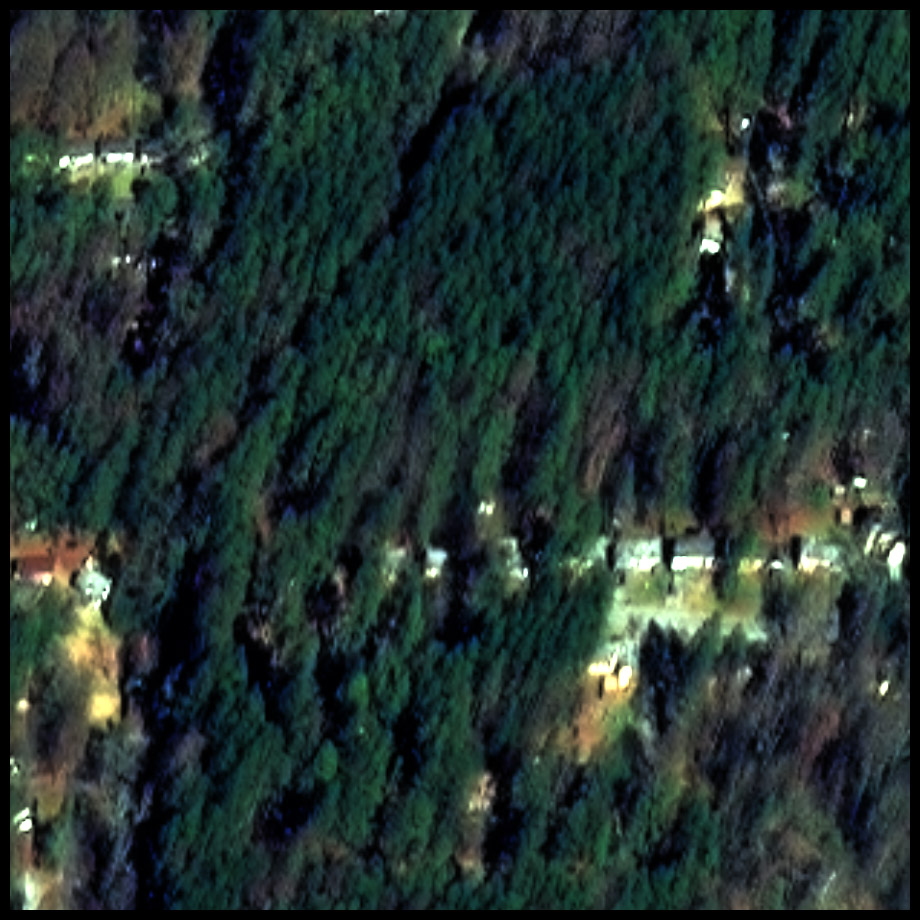}} \\
\end{tabular}
\caption{\textbf{Sample imagery from SpaceNet MVOI.} Four of the 2222 geographically unique image chips in the dataset are shown (columns), with three of the 27 views of that chip (rows), one from each angle bin. Negative look angle corresponds to South-facing views, whereas positive look angles correspond to North-facing views (Figure \ref{collect_angle_figure}). Chips are down-sampled from $900\times900$ pixel high-resolution images. In addition to the RGB images shown, the dataset comprises a high-resolution pan-chromatic (grayscale) band, a high-resolution near-infrared band, and a lower-resolution 8-band multispectral image for each geographic location/view combination. The dataset is available at \url{https://spacenet.ai} under a CC-BY SA 4.0 License.}
\label{fig:ex_comic_strip1}
\end{center}
\vspace{-20pt}
\end{figure*}

Though public overhead imagery datasets explore geographic and sensor homogeneity \cite{DeepGlobe18, Etten2018SpaceNetAR, Veryhigh-BaseCollect, Xia:2017, Lam:2018}, they generally comprise a single view of the imaged location(s) taken nearly directly overhead (``at nadir"). Nadir imagery is not representative of collections during disaster response or other urgent situations: for example, the first public high-resolution cloud-free image of San Juan, Puerto Rico following Hurricane Maria was taken at $51.9^\circ$ ``off-nadir", \textit{i.e.}, a $51.9^\circ$ angle between the nadir point directly underneath the satellite and the center of the imaged scene \cite{DGdiscover}. The disparity between looks in public training data and relevant use cases hinders development of models applicable to real-world problems. More generally, satellite and drone images rarely capture identical looks at objects in different contexts, or even when repeatedly imaging the same geography. Furthermore, no existing datasets or metrics permit assessment of model robustness to different looks, prohibiting evaluation of performance. These limitations extend to tasks outside of the geospatial domain: for example, convolutional neural nets perform inconsistently in many natural scene video frame classification tasks despite minimal pixel-level variation \cite{azulay2018}, and Xiao et al. showed that spatial transformation of images, effectively altering view, represents an effective adversarial attack against computer vision models \cite{xiao2018}. Addressing generalization across views both within and outside of the geospatial domain requires two advancements: 1. A large multi-view dataset with diversity in land usage, population density, and views, and 2. A metric to assess model generalization.

To address the limitations detailed above, we introduce the SpaceNet Multi-View Overhead Imagery (MVOI) dataset, which includes 62,000 overhead images collected over Atlanta, Georgia USA and the surrounding areas. The dataset comprises 27 distinct looks, including both North- and South-facing views, taken during a single pass of a Maxar WorldView-2 satellite. The looks range from almost directly overhead ($7.8^\circ$ off-nadir) to up to $54^\circ$ off-nadir, with the same $665\ km^2$ geographic area covered by each. Alongside the imagery we open sourced an attendant 126,747 building footprints created by expert labelers. To our knowledge, this is the first multi-viewpoint dataset for overhead imagery with dense object annotations. The dataset covers heterogeneous geographies, including highly treed rural areas, suburbs, industrial areas, and high-density urban environments, resulting in heterogeneous building size, density, context and appearance (Figure \ref{fig:ex_comic_strip1}). At the same time, the dataset abstracts away many other time-sensitive variables (\textit{e.g.} seasonality), enabling careful assessment of the impact of look angle on model training and inference. The training imagery and labels and public test images are available at https://spacenet.ai under the CC-BY SA 4.0 International License.

Though an ideal overhead imagery dataset would cover all the variables present in overhead imagery, \textit{i.e.} look angle, seasonality, geography, weather condition, sensor, and light conditions, creating such a dataset is impossible with existing imagery. To our knowledge, the 27 unique looks in SpaceNet MVOI represent one of only two such imagery collections available in the commercial realm, even behind imagery acquisition company paywalls. We thus chose to focus SpaceNet MVOI on providing a diverse set of views with varying look angle and direction, a variable that is not represented in any existing overhead imagery dataset. SpaceNet MVOI could potentially be combined with existing datasets to train models which generalize across more variables.

We benchmark state-of-the art models on three tasks:
\begin{enumerate}[noitemsep]
    \item Building segmentation and detection.
    \item Generalization of segmentation and object detection models to previously unseen angles.
    \item Consequences of changes in resolution for segmentation and object detection models.
\end{enumerate}

Our benchmarking reveals that state-of-the-art detectors are challenged by SpaceNet MVOI, particularly in views left out during model training. Segmentation and object detection models struggled to account for displacement of building footprints, occlusion, shadows, and distortion in highly off-nadir looks (Figure \ref{fig:challenges}). The challenge of addressing footprint displacement is of particular interest, as it requires models not only to learn visual features, but to adjust footprint localization dependent upon the view context. Addressing these challenges is relevant to a number of applications outside of overhead imagery analysis, \textit{e.g.} autonomous vehicle vision.

To assess model generalization to new looks we developed a generalization metric $G$, which reports the relative performance of models when they are applied to previously unseen looks. While specialized models designed for overhead imagery out-perform general baseline models in building footprint detection, we found that models developed for natural image computer vision tasks have better $G$ scores on views absent during training. These observations highlight the challenges associated with developing robust models for multi-view object detection and semantic segmentation tasks. We therefore expect that developments in computer vision models for multi-view analysis made using SpaceNet MVOI, as well as analysis using our metric $G$, will be broadly relevant for many computer vision tasks.

The dataset is available at \url{www.spacenet.ai}.


\section{Related Work}

Object detection and segmentation is a well-studied problem for natural scene images, but those objects are generally much larger and suffer minimally from distortions exacerbated in overhead imagery. Natural scene research is driven by datasets such as MSCOCO \cite{coco} and PASCALVOC \cite{pascalvoc}, but those datasets lack multiple views of each object. PASCAL3D \cite{pascal3D}, autonomous driving datasets such as KITTI \cite{kitti}, CityScapes \cite{Cordts_2016_CVPR}, existing multi-view datasets \cite{Simon_2017, Sridhar2013InteractiveMA}, and tracking datasets such as MOT2017\cite{MOT16} or OBT \cite{obt} contains different views but are confined to a narrow range of angles, lack sufficient heterogeneity to test generalization between views, and are restricted to natural scene images. Multiple viewpoints are found in 3D model datasets \cite{3dmodels,lotter}, but those are not photorealistic and lack the occlusion and visual distortion properties encountered with real imagery.

Previous datasets for overhead imagery focus on classification \cite{fmow}, bounding box object detection \cite{Xia:2017, Lam:2018, Mundhenk:2016}, instance-based segmentation \cite{Etten2018SpaceNetAR}, and object tracking \cite{Robicquet} tasks. None of these datasets comprise multiple images of the same field of view from substantially different look angles, making it difficult to assess model robustness to new views. Within segmentation datasets, SpaceNet \cite{Etten2018SpaceNetAR} represents the closest work, with dense building and road annotations created by the same methodology. We summarize the key characteristics of each dataset in Table \ref{datasets}. Our dataset matches or exceeds existing datasets in terms of imagery size and annotation density, but critically includes varying look direction and angle to better reflect the visual heterogeneity of real-world imagery.

\begin{figure}[t]
\vspace{-20pt}
\begin{center}
\setlength{\tabcolsep}{0.15em}
\begin{tabular}{ccc}
\vspace{-5pt}
& \multicolumn{2}{c}{\textbf{Challenges in off-nadir imagery}} \\
\vspace{-5pt}
\raisebox{0.63in}{
   \rotatebox[origin=c]{90}{\makecell{\textbf{Footprint offset} \\ \textbf{and occlusion}}}} &
\subfloat[\textbf{7 degrees}]{\includegraphics[width=0.42\linewidth]{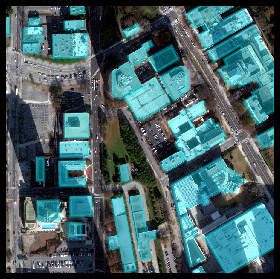}} &
\subfloat[\textbf{53 degrees}]{\includegraphics[width=0.42\linewidth]{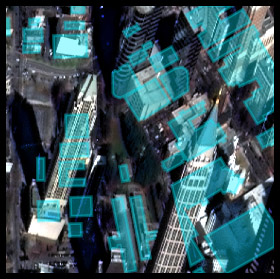}}  \\

\vspace{-5pt}
\raisebox{0.63in}{\rotatebox[origin=c]{90}{\textbf{Shadows}}} &
\subfloat[\textbf{30 degrees}]{\includegraphics[width=0.42\linewidth]{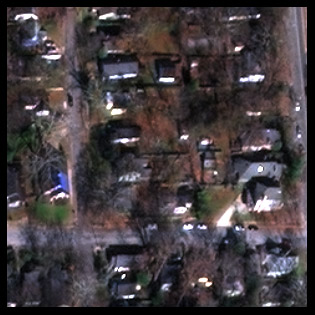}} &
\subfloat[\textbf{-32 degrees}]{\includegraphics[width=0.42\linewidth]{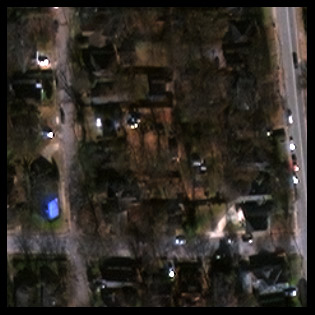}}  \\
\end{tabular}

\caption{\textbf{Challenges with off-nadir look angles.} Though geospatially accurate building footprints (blue) perfectly match building roofs at nadir \textit{\textbf{(A)}}, this is not the case off-nadir \textit{\textbf{(B)}}, and many buildings are obscured by skyscrapers. \textit{\textbf{(C-D):}} Visibility of some buildings changes at different look angles due to variation in reflected sunlight.}
\label{fig:challenges}
\end{center}
\vspace{-20pt}
\end{figure}

The effect of different views on segmentation or object detection in natural scenes has not been thoroughly studied, as feature characteristics are relatively preserved even under rotation of the object in that context. Nonetheless, preliminary studies of classification model performance on video frames suggests that minimal pixel-level changes can impact performance \cite{azulay2018}. By contrast, substantial occlusion and distortion occurs in off-nadir overhead imagery, complicating segmentation and placement of geospatially accurate object footprints, as shown in Figure \ref{fig:challenges}A-B. Furthermore, due to the comparatively small size of target objects (\textit{e.g.} buildings) in overhead imagery, changing view substantially alters their appearance (Figure \ref{fig:challenges}C-D). We expect similar challenges to occur when detecting objects in natural scene images at a distance or in crowded views. Existing solutions to occlusion are often domain specific \cite{Yang_2015} or rely on attention mechanisms to identify common elements \cite{Zhang_2018_CVPR} or landmarks \cite{Yuen2017AnOS}. The heterogeneity in building appearance in overhead imagery, and the absence of landmark features to identify them, makes their detection an ideal research task for developing domain-agnostic models that are robust to occlusion.

\begin{table*}[bt]
\begin{center}
\begin{tabular}{lrrrrrr}
\textbf{Dataset} & \textbf{Gigapixels} & \textbf{\# Images} & \textbf{Resolution (m)} & \textbf{Nadir Angles} & \textbf{\# Objects} & \textbf{Annotation}\\
\hline
\hline
SpaceNet \cite{Etten2018SpaceNetAR, DeepGlobe18} & 10.3 & 24586 & 0.31 & On-Nadir & 302701 & Polygons \\
DOTA \cite{Xia:2017}  & 44.9 & 2806 & Google Earth* & On-Nadir & 188282 & Oriented Bbox\\
3K Vehicle Detection \cite{FastMV}& N/A & 20 & 0.20 & Aerial & 14235 & Oriented Bbox\\
UCAS-AOD \cite{ucas} & N/A & 1510 & Google Earth* & On-Nadir & 3651 & Oriented Bbox \\
NWPU VHR-10 \cite{LearningRC} & N/A & 800 & Google Earth* & On-Nadir & 3651 & Bbox \\
MVS \cite{MVS} & 111 & 50 & 0.31-0.58 & {[}5.3, 43.3{]} & 0 & None \\
FMoW \cite{fmow} & 1,084.0 & 523846 & 0.31-1.60 & {[}0.22, 57.5{]} & 132716 & Classification \\
xView \cite{Lam:2018} & 56.0 & 1400 & 0.31 & On-Nadir & $1000000$ & Bbox\\
\hline
\textbf{SpaceNet MVOI (Ours)} & \textbf{50.2} & \textbf{60000} & \textbf{0.46-1.67} & \textbf{{[}-32.5, +54.0{]}} & \textbf{126747} & \textbf{Polygons}  \\
\hline
PascalVOC \cite{pascalvoc} & - & 21503 & - & - & 62199 & Bbox \\
MSCOCO \cite{coco} & - & 123287 & - & - & 886266 & Bbox \\
ImageNet \cite{imagenet_cvpr09} & - & 349319 & - & - & 478806 & Bbox \\
\end{tabular}
\end{center}
\vspace{-10pt}
\caption{\textbf{Comparison with other computer vision and overhead imagery datasets.} Our dataset has a similar scale as modern computer vision datasets, but to our knowledge is the first multi-view overhead imagery dataset designed for segmentation and object detection tasks. *Google Earth imagery is a mosaic from a variety of aerial and satellite sources and ranges from 15 cm to 12 m resolution \cite{googleearth}.}
\label{datasets}
\vspace{-10pt}
\end{table*}

\section{Dataset Creation}

\begin{figure}[tb]
    \centering
    \includegraphics[width=0.5\textwidth]{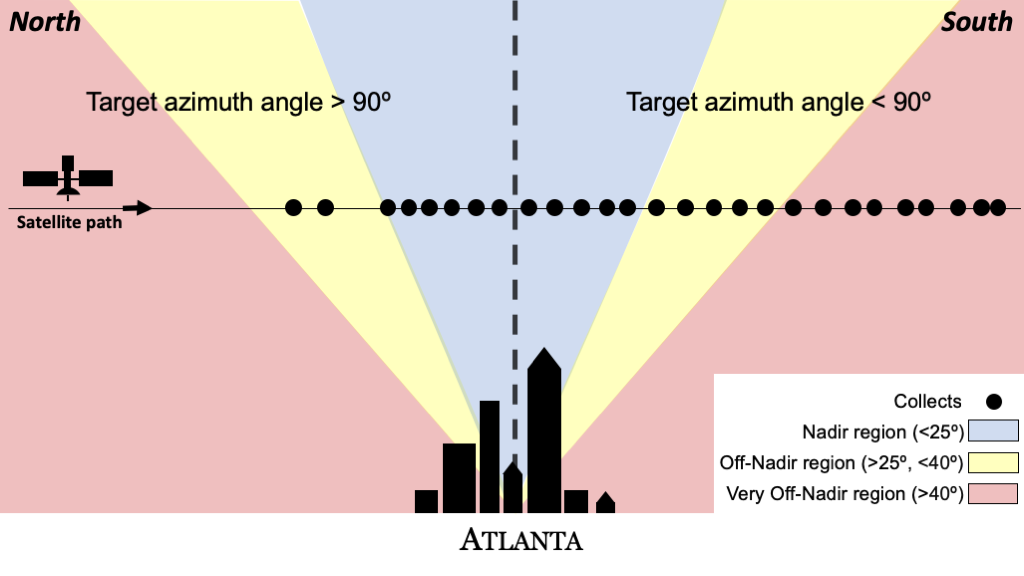}
    \caption{\textbf{Collect views.} Location of collection points during the WorldView-2 satellite pass over Atlanta, GA USA.}
    \label{collect_angle_figure}
    \vspace{-10pt}
\end{figure}

SpaceNet MVOI contains images of Atlanta, GA USA and surrounding geography collected by Maxar\textquoteright s WorldView-2 Satellite on December 22, 2009 \cite{Veryhigh-BaseCollect}. The satellite collected 27 distinct views of the same 665 km\textsuperscript{2} ground area during a single pass over a 5 minute span. This produced 27 views with look angles (angular distance between the nadir point directly underneath the satellite and the center of the scene) from $7.8^\circ$ to $54^\circ$ off-nadir and with a target azimuth angle (compass direction of image acquisition) of $17^\circ$ to $182.8^\circ$ from true North (see Figure \ref{collect_angle_figure}).  See the Supplementary Material and Tables S1 and S2 for further details regarding the collections. The 27 views in a narrow temporal band provide a dense set of visually distinct perspectives of static objects (buildings, roads, trees, utilities, etc.) while limiting complicating factors common to remote sensing datasets such as changes in cloud cover, sun angle, or land-use change. The imaged area is geographically diverse, including urban areas, industrial zones, forested suburbs, and undeveloped areas (Figure \ref{fig:ex_comic_strip1}).

\subsection{Preprocessing}
Multi-view satellite imagery datasets are distinct from related natural image datasets in several interesting ways. First, as look angle increases in satellite imagery, the native resolution of the image decreases because greater distortion is required to project the image onto a flat grid (Figure \ref{fig:ex_comic_strip1}). Second, each view contains images with multiple spectral bands. For the purposes our baselines, we used 3-channel images (RGB: red, green, blue), but also examined the contributions of the near-infrared (NIR) channel (see Supplementary Material). These images were enhanced with a separate, higher resolution panchromatic (grayscale) channel to double the original resolution of the multispectral imagery (\textit{i.e.}, ``pan-sharpened"). The entire dataset was tiled into $900px \times 900px$ tiles and resampled to simulate a consistent resolution across all viewing angles of $0.5m \times 0.5m$ ground sample distance. The dataset also includes lower-resolution 8-band multispectral imagery with additional color channels, as well as panchromatic images, both of which are common overhead imagery data types.

The 16-bit pan-sharpened RGB-NIR pixel intensities were truncated at 3000 and then rescaled to an 8-bit range before normalizing to $[0, 1]$. We also trained models directly using Z-score normalized 16 bit images with no appreciable difference in the results.

\subsection{Annotations}
We undertook professional labeling to produce high-quality annotations. An expert geospatial team exhaustively labeled building footprints across the imaged area using the most on-nadir image ($7.8^\circ$ off-nadir). Importantly, the building footprint polygons represent geospatially accurate ground truth, and therefore are shared across all views. For structures occluded by trees, only the visible portion was labeled. Finally, one independent validator and one remote sensing expert evaluated the quality of each label.

\begin{figure}[tb]
    \centering
    \includegraphics[width=0.5\textwidth]{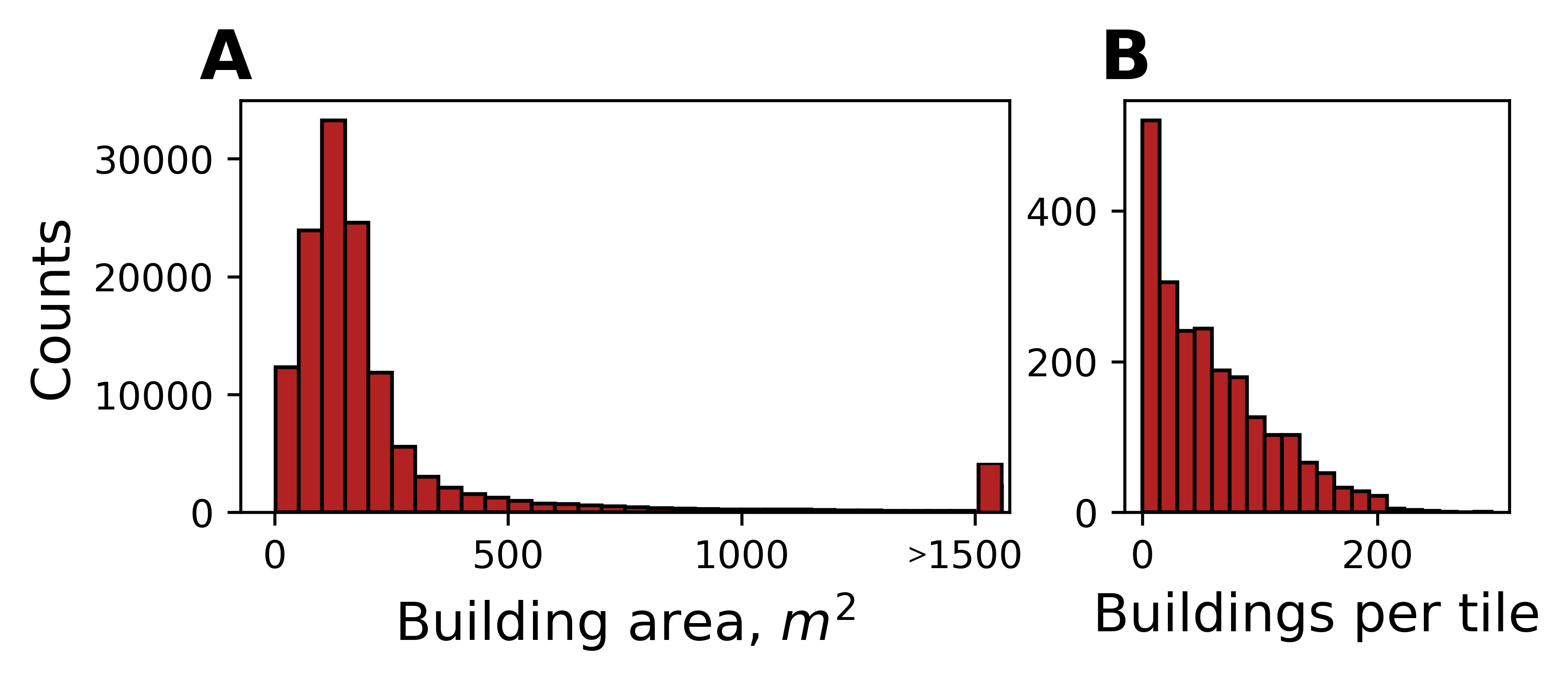}
    \vspace{-10pt}
    \caption{\textbf{Dataset statistics.} Distribution of \textbf{(A)} building footprint areas and \textbf{(B)} number of objects per $450 m \times 450 m$ geographic tile in the dataset.}
    \label{stats}
\vspace{-15pt}
\end{figure}

\subsection{Dataset statistics}

Our dataset labels comprise a broad distribution of building sizes, as shown in Figure \ref{stats}A. Compared to natural image datasets, our dataset more heavily emphasizes small objects, with the majority of objects less than 700 pixels in area, or $\sim25$ pixels across. By contrast, objects in the PASCALVOC \cite{pascalvoc} or MSCOCO \cite{coco} datasets usually comprise 50-300 pixels along the major axis \cite{Xia:2017}.

An additional challenge presented by this dataset, consistent with many real-world computer vision tasks, is the heterogeneity in target object density (Figure \ref{stats}B). Images contained between zero and 300 footprints, with substantial coverage throughout that range. This variability presents a challenge to object detection algorithms, which often require estimation of the number of features per image \cite{He_2017}. Segmentation and object detection of dense or variable density objects is challenging, making this an ideal dataset to test the limits of algorithms' performance.

\begin{table}[bt]
\begin{center}
\begin{tabular}{ll}
\textbf{Task} & \textbf{Baseline models} \\
\hline
\hline
Semantic Segmentation & TernausNet \cite{ternausnet} , U-NET \cite{Ronneberger:2015gk} \\
 Instance Segmentation & Mask R-CNN \cite{He_2017}\\
Object Detection & Mask R-CNN \cite{He_2017}, YOLT \cite{yolt} \\
\end{tabular}
\vspace{-10pt}
\end{center}
\caption{Benchmark model selections for dataset baselines. TernausNet and YOLT are overhead imagery-specific models, whereas Mask R-CNN and U-Net are popular natural scene analysis models.}
\label{models}
\vspace{-10pt}
\end{table}

\section{Building Detection Experiments}

\subsection{Dataset preparation for analysis}

We split the training and test sets 80/20 by randomly selecting geographic locations and including all views for that location in one split, ensuring that each type of geography was represented in both splits. We group each angle into one of three categories: Nadir (NADIR), $\theta \leq  25^\circ$; Off-nadir (OFF), $25^\circ < \theta < 40^\circ$; and Very off-nadir (VOFF), $\theta \geq 40^\circ$.
In all experiments, we trained baselines using all viewing angles (ALL) or one of the three subsets. These trained models were then evaluated on the test set of each of the 27 viewing angles individually.

\subsection{Models}

We measured several state of the art baselines for semantic or instance segmentation and object detection (Table \ref{models}). Where possible, we selected overhead imagery-specific models as well as models for natural scenes to compare their performance. Object detection baselines were trained using rectangular boundaries extracted from the building footprints. To fairly compare with semantic segmentation studies, the resulting bounding boxes were compared against the ground truth building polygons for scoring (see Metrics).

\subsection{Segmentation Loss}

Due to the class imbalance of the training data -- only 9.5\% of the pixels in the training set correspond to buildings -- segmentation models trained with binary cross-entropy (BCE) loss failed to identify building pixels, a problem observed previously for overhead imagery segmentation models \cite{Sun_2018_CVPR_Workshops}. For the semantic segmentation models, we therefore utilized a hybrid loss function that combines the binary cross entropy loss and intersection over union (IoU) loss with a weight factor $\alpha$ \cite{Sun_2018_CVPR_Workshops}:

\begin{equation} \label{eq:1}
    \mathcal{L} = \alpha\mathcal{L}_{\text{BCE}} + (1-\alpha)\mathcal{L}_{\text{IoU}}
\end{equation}

\noindent The details of model training and evaluation, including augmentation, optimizers, and evaluation schemes can be found in the Supplementary Material.

\subsection{Metrics}

We measured performance using the building IoU-$F_1$ score defined in Van Etten et al. \cite{Etten2018SpaceNetAR}. Briefly, building footprint polygons were extracted from segmentation masks (or taken directly from object detection bounding box outputs) and compared to ground truth polygons. Predictions were labeled True Positive if they had an IoU with a ground truth polygon above 0.5 and all other predictions were deemed False Positives. Using these statistics and the number of undetected ground truth polygons (False Negatives), we calculated the precision $P$ and recall $R$ of the model predictions in aggregate. We then report the $F_1$ score as

\begin{equation}
    F_1 = \frac{2 \times P \times R}{P + R}
\end{equation}

$F_1$ score was calculated within each angle bin (NADIR, OFF, or VOFF) and then averaged for an aggregate score.

\begin{table}[bt]
\vspace{-10pt}
\begin{center}
\begin{tabular}{llrrrr}
             &          & \multicolumn{4}{c}{$\boldsymbol{F_1}$} \\
\textbf{Task}& \textbf{Model} & NADIR   & OFF    & VOFF & Avg. \\
\hline
\hline
Seg     & TernausNet & \textbf{0.62}    & \textbf{0.43}   & \textbf{0.22} & \textbf{0.43} \\
Seg     & U-Net         & 0.39    & 0.27   & 0.08 & 0.24  \\
Seg     & Mask R-CNN     & 0.47    & 0.34   & 0.07 & 0.29 \\
\hline
Det & Mask R-CNN     & 0.40    & 0.30   & 0.07 & 0.25 \\
Det & YOLT          & 0.49    & 0.37   & 0.20 & 0.36 \\
\end{tabular}
\end{center}
\vspace{-10pt}
\caption{\textbf{Overall task difficulty.} As a measure of overall task difficulty, the performance ($F_1$ score) is assessed for the baseline models trained on all angles, and tested on the three different viewing angle bins: nadir (NADIR), off-nadir (OFF), and very off-nadir (VOFF). Avg. is the linear mean of the three bins. Seg, segmentation; Det, object detection.}
\label{trainall}
\vspace{-10pt}
\end{table}

\subsection{Results}


\begin{figure*}[t]
\setlength{\tabcolsep}{0.2em}
\begin{center}
\begin{tabular}{cccccc}
\vspace{-10pt}
 & & \textbf{Image} & \textbf{Mask R-CNN} & \textbf{TernausNet} & \textbf{YOLT} \\

\vspace{-10pt}

&
\raisebox{0.625in}{\rotatebox[origin=c]{90}{10 (NADIR)}} &
\subfloat{\includegraphics[width=0.18\linewidth]{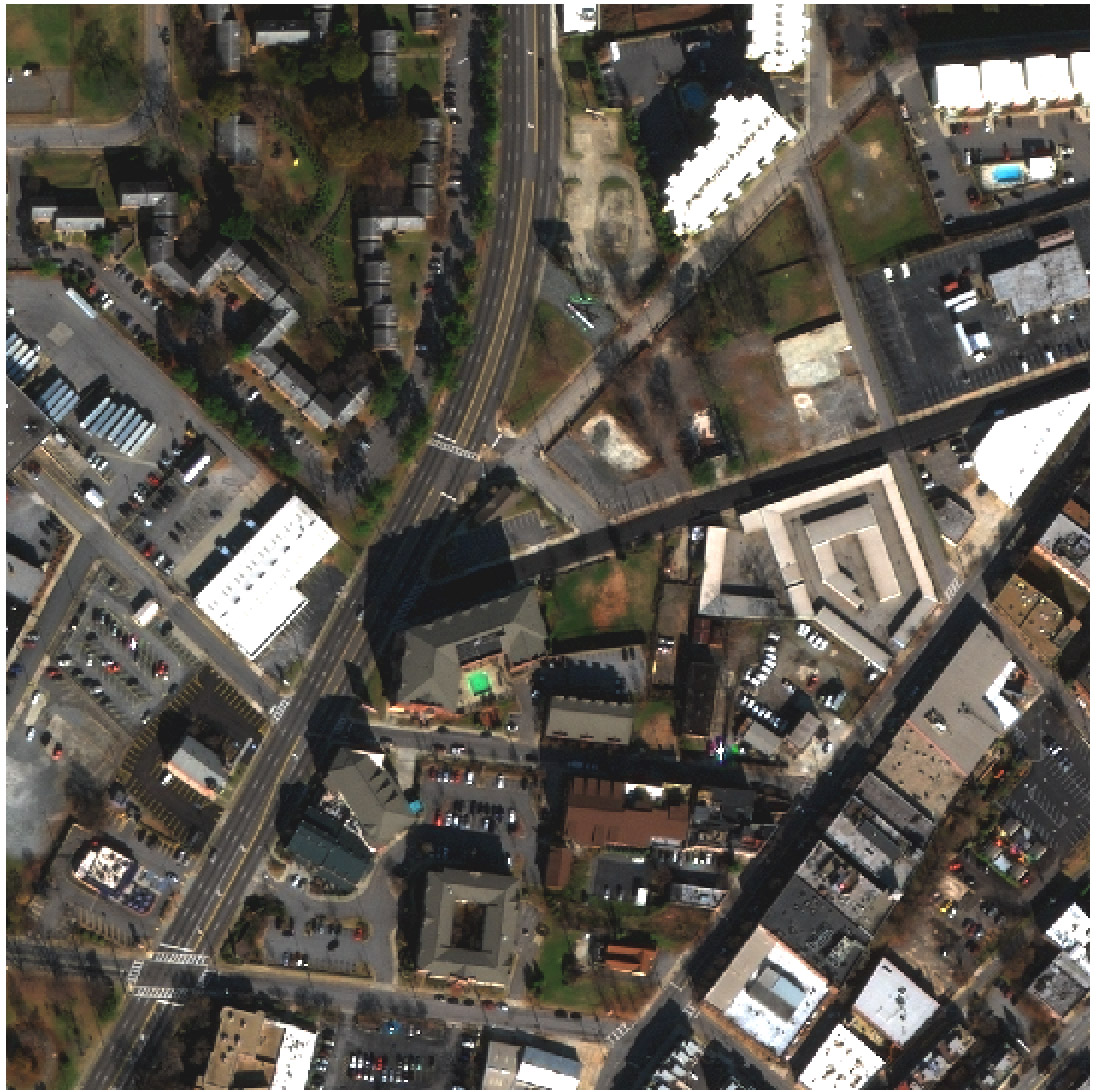}} &
\subfloat{\includegraphics[width=0.18\linewidth]{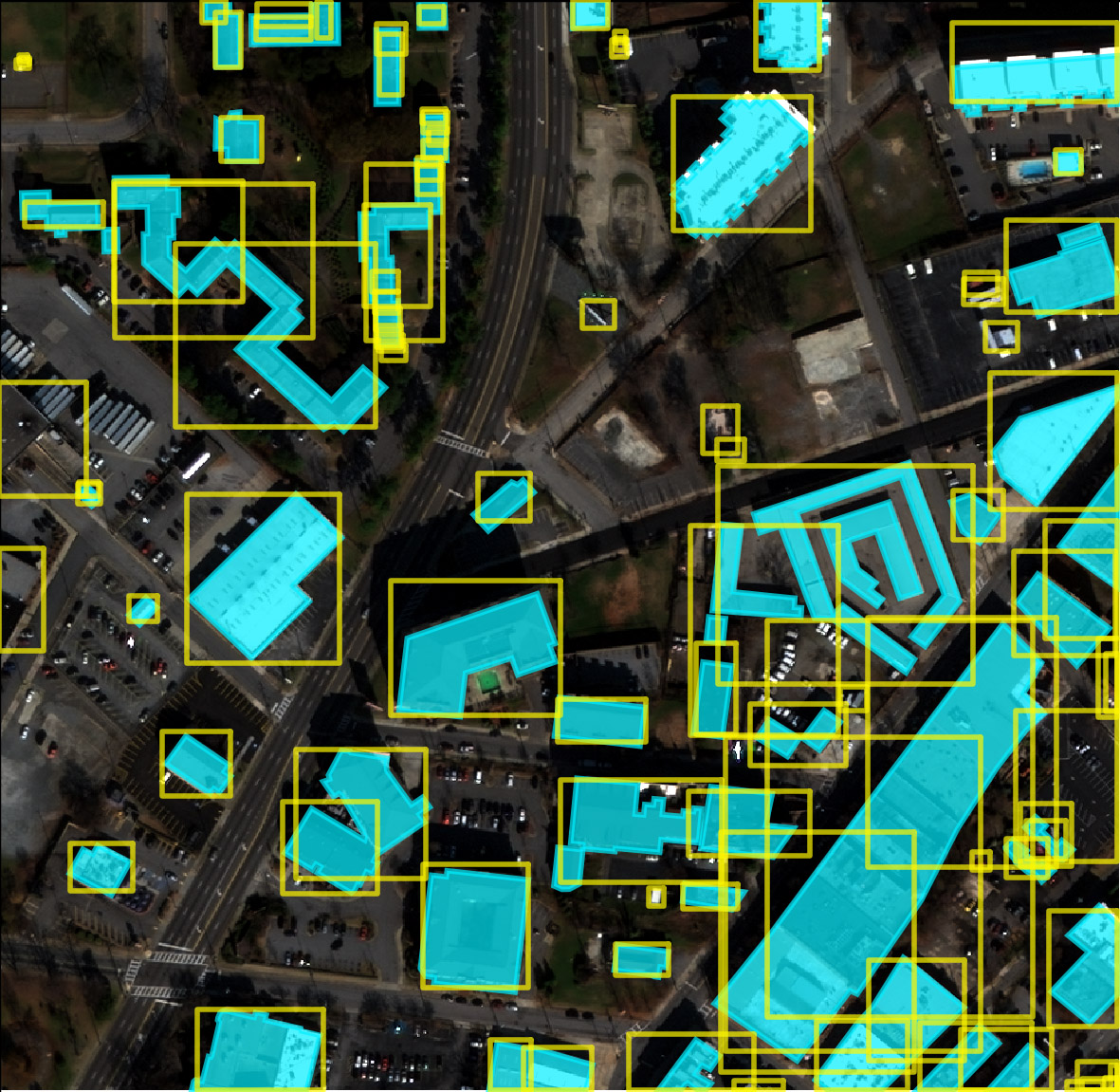}} &
\subfloat{\includegraphics[width=0.18\linewidth]{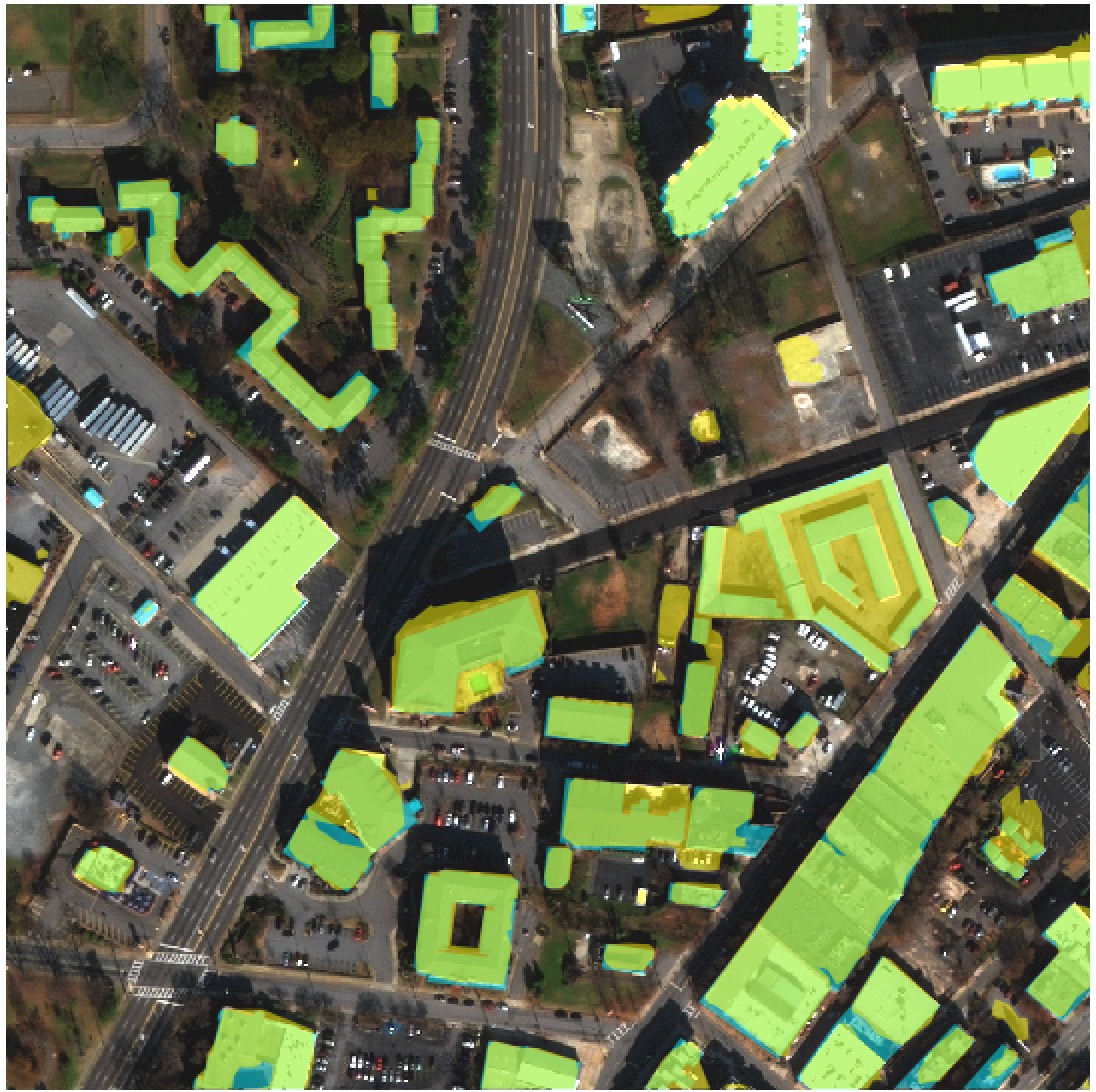}} &
\subfloat{\includegraphics[width=0.18\linewidth]{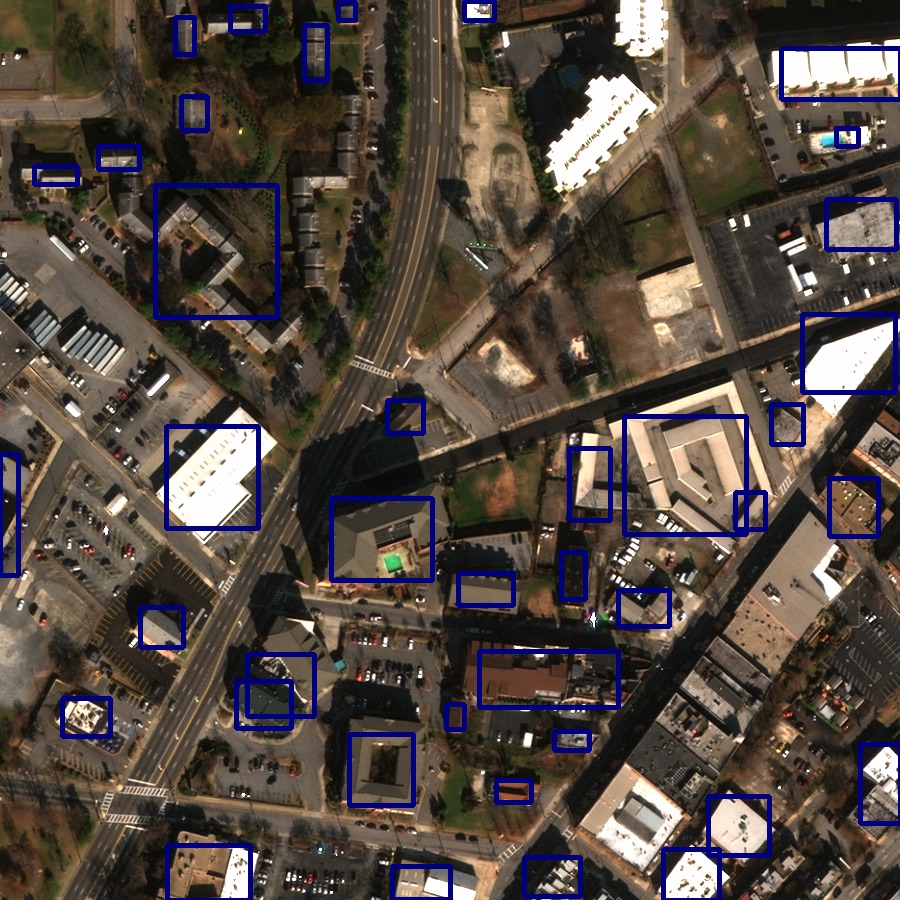}} \\

\vspace{-10pt}

\raisebox{0.63in}{
   \rotatebox[origin=c]{90}{\makecell{\textbf{LOOK ANGLE} \\ \textbf{(BIN)}}}} & \raisebox{0.625in}{
   \rotatebox[origin=c]{90}{-29 (OFF)}} &
\subfloat{\includegraphics[width=0.18\linewidth]{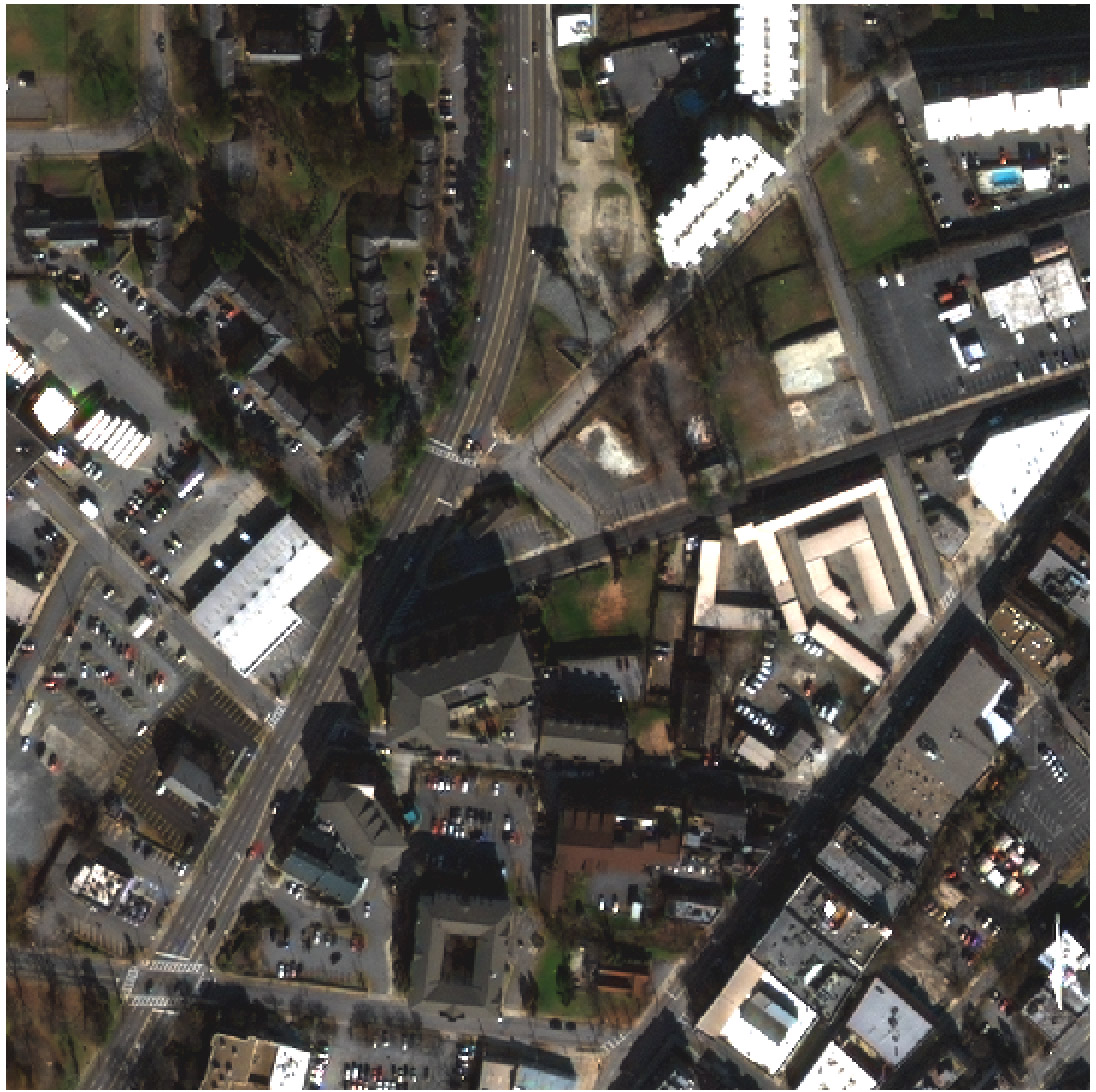}} &
\subfloat{\includegraphics[width=0.18\linewidth]{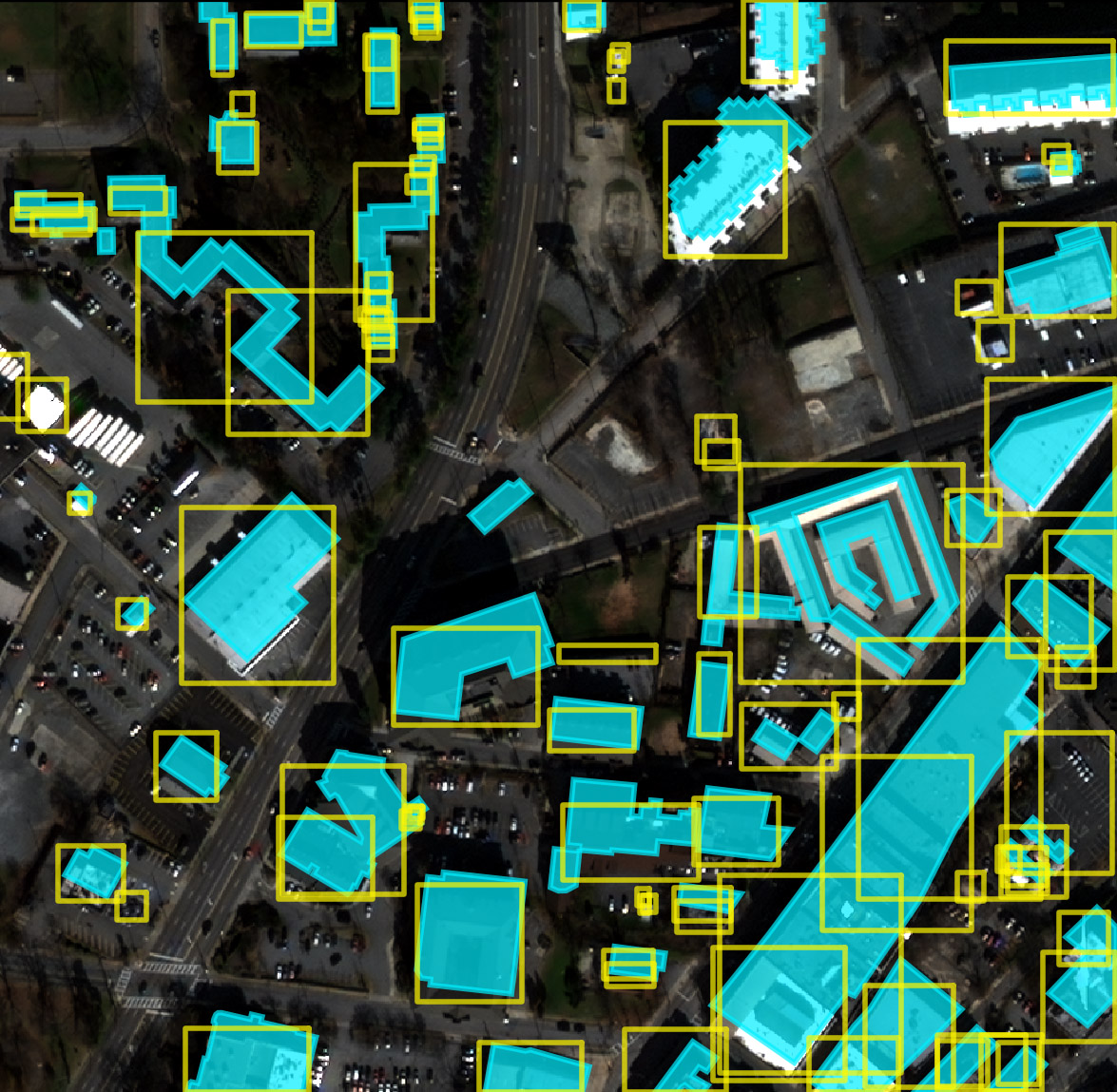}} &
\subfloat{\includegraphics[width=0.18\linewidth]{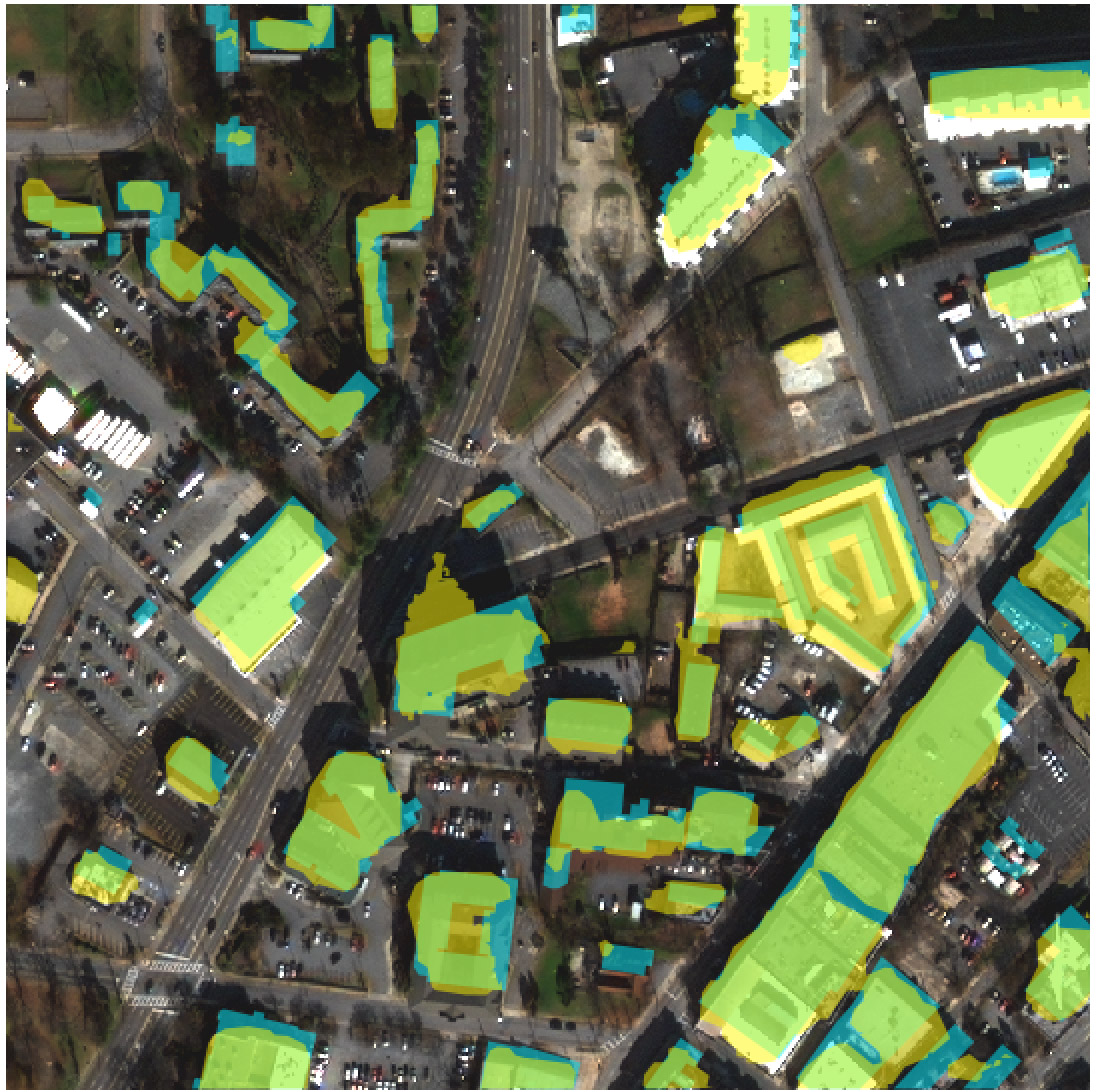}} &
\subfloat{\includegraphics[width=0.18\linewidth]{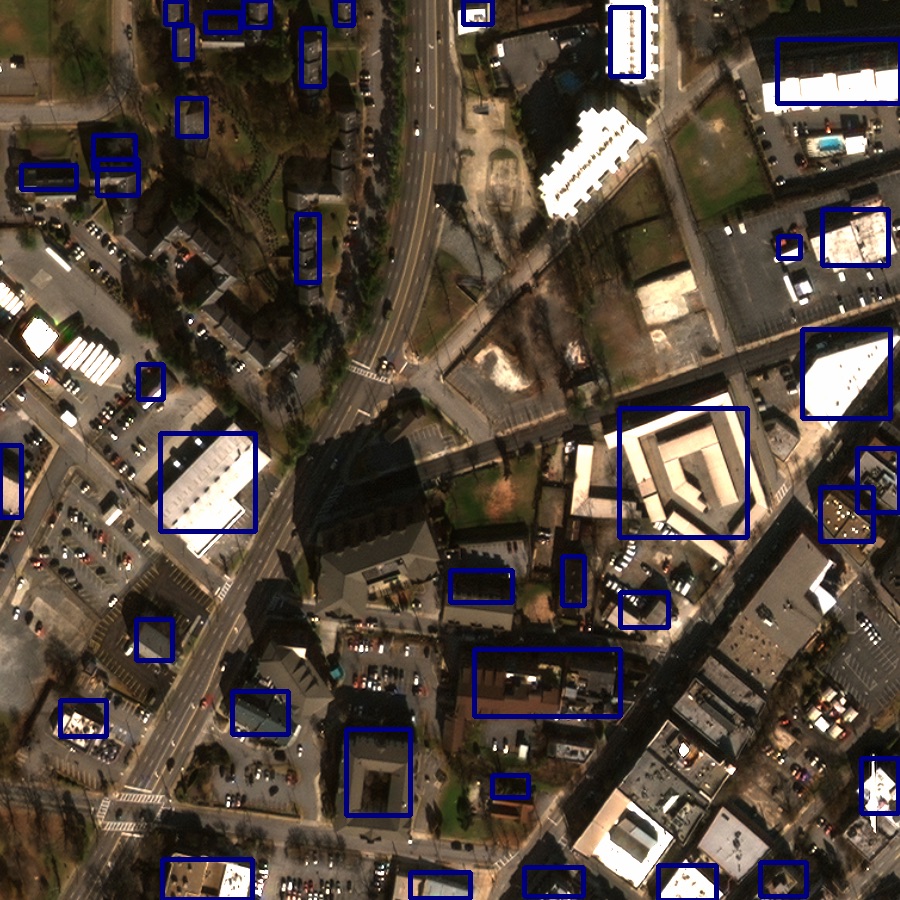}} \\

\vspace{-10pt}

& \raisebox{0.625in}{
   \rotatebox[origin=c]{90}{53 (VOFF)}} &
\subfloat{\includegraphics[width=0.18\linewidth]{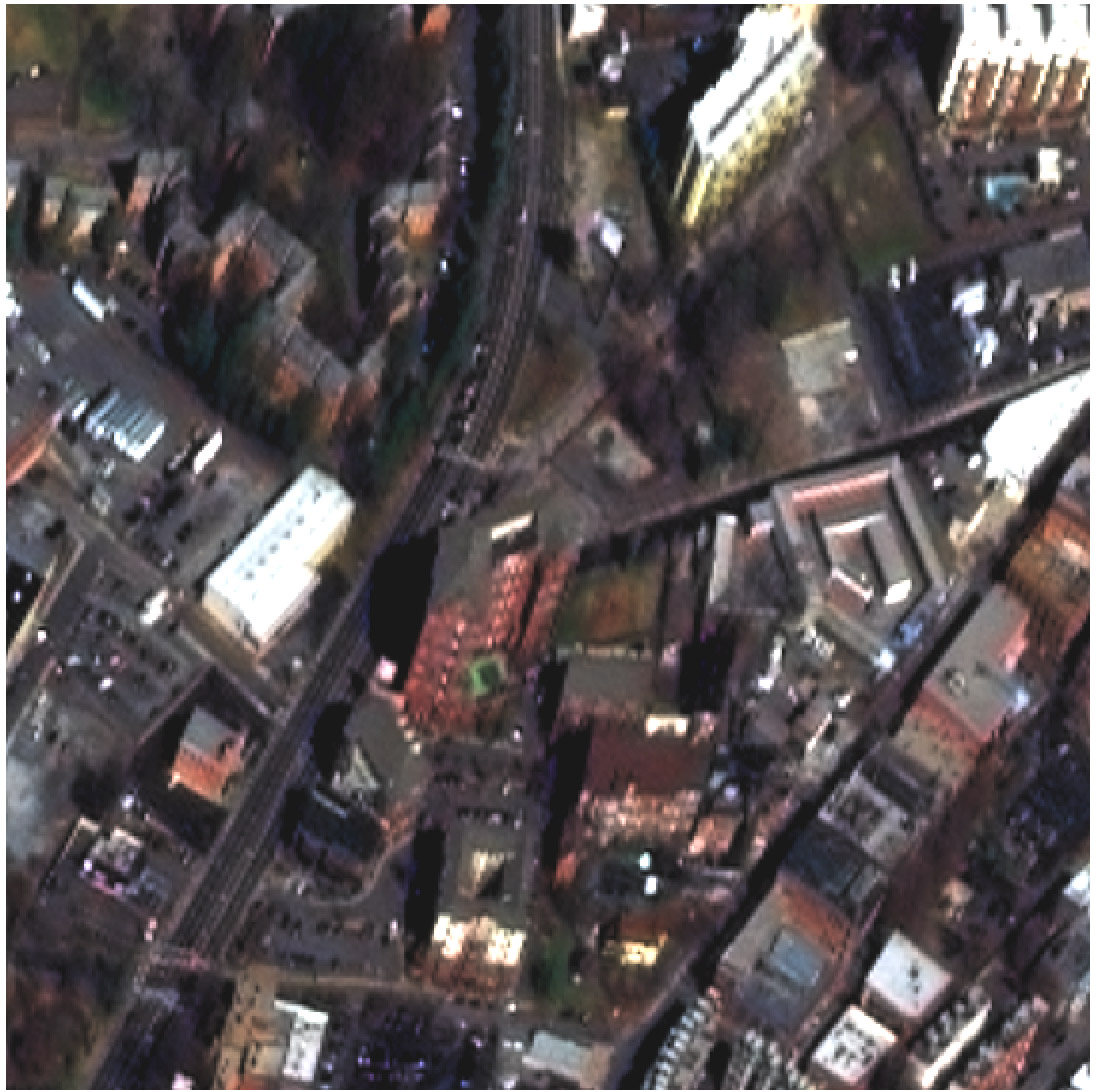}} &
\subfloat{\includegraphics[width=0.18\linewidth]{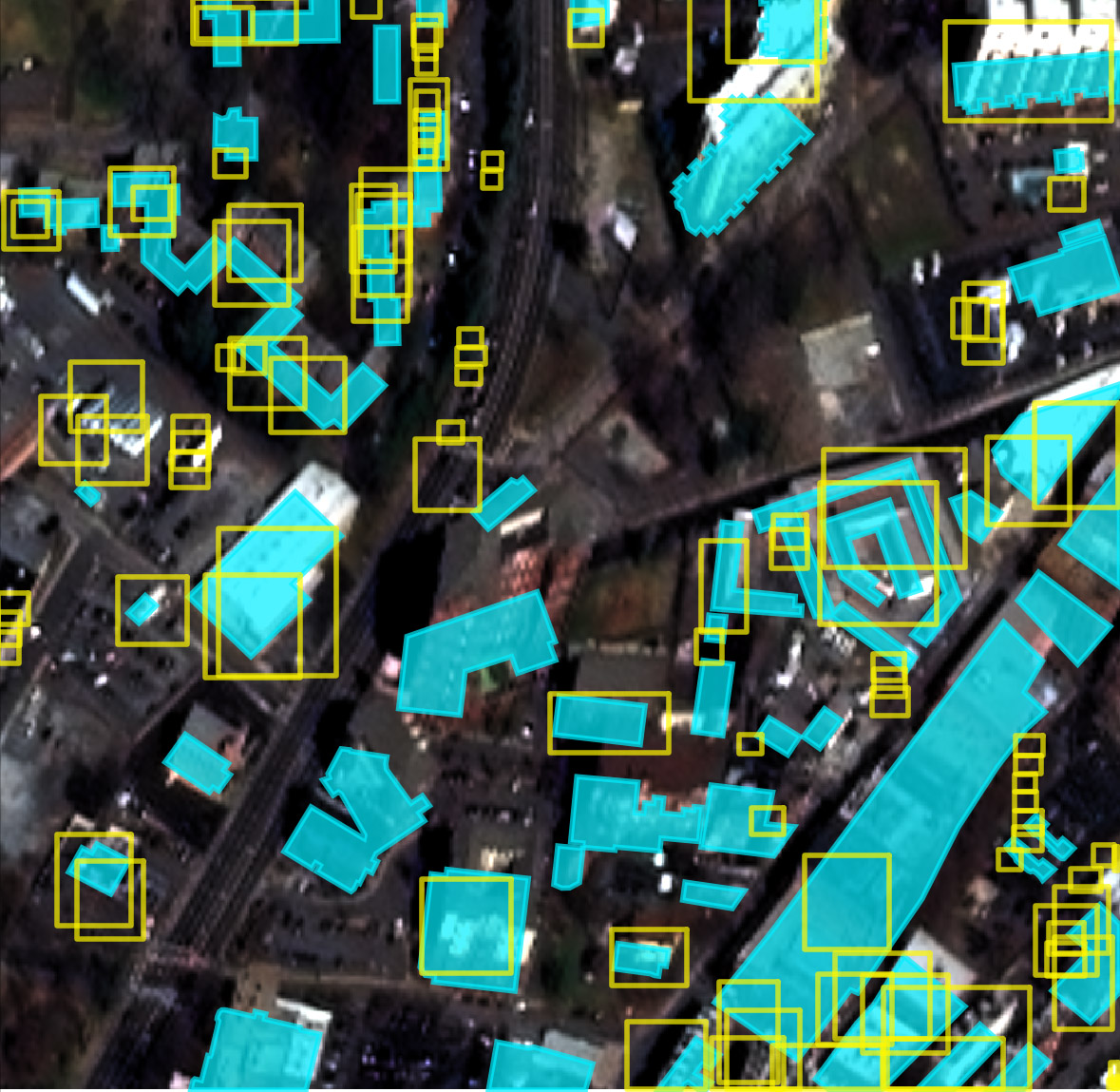}} &
\subfloat{\includegraphics[width=0.18\linewidth]{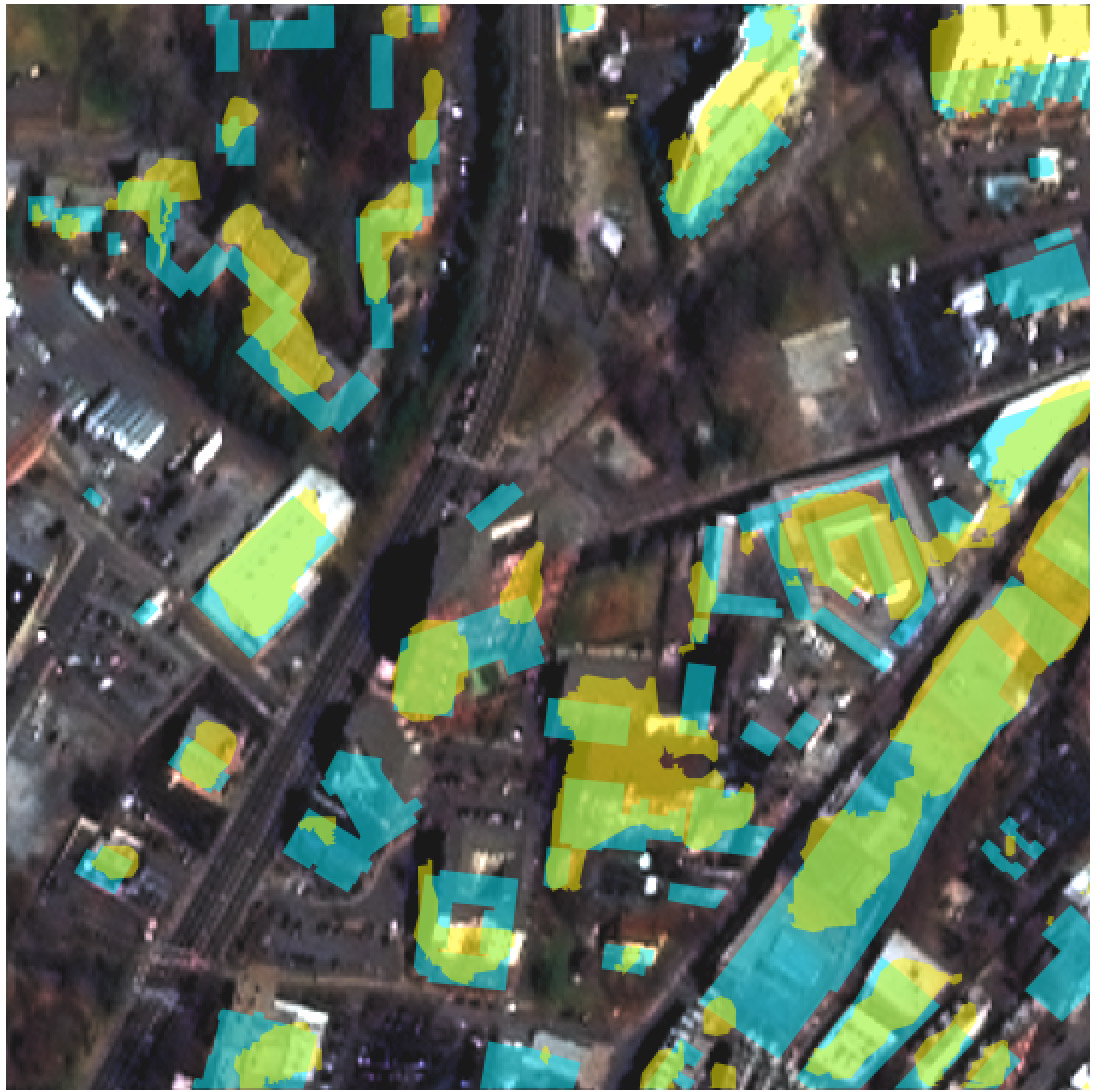}} &
\subfloat{\includegraphics[width=0.18\linewidth]{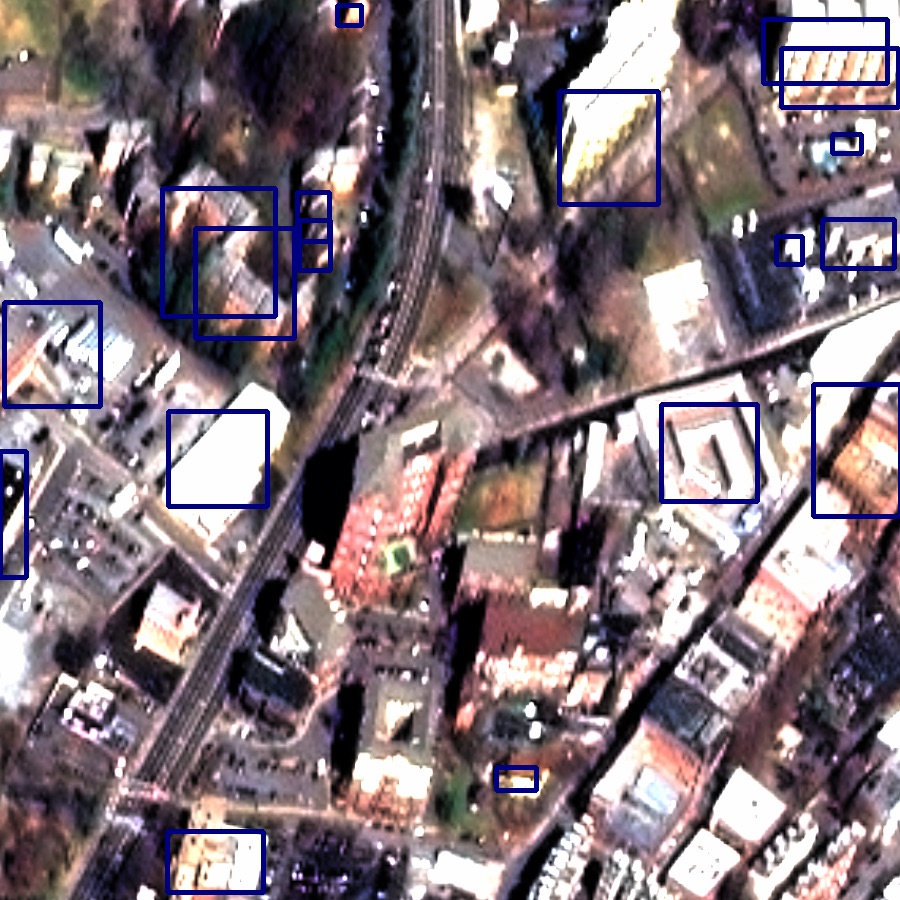}} \\

\end{tabular}
 \caption{\textbf{Sample imagery (left) with ground truth building footprints and Mask R-CNN bounding boxes (middle left), TernausNet segmentation masks (middle right), and YOLT bounding boxes (right).} Ground truth masks (light blue) are shown under Mask R-CNN and TernausNet predictions (yellow). YOLT bounding boxes shown in blue. Sign of the look angle represents look direction (negative=south-facing, positive=north-facing). Predictions from models trained on on all angles (see Table \ref{trainall}). }
\label{fig:comic_strip}
\end{center}
\vspace{-25pt}
\end{figure*}

\begin{figure*}[tb]
    \centering
    \includegraphics[width=\textwidth]{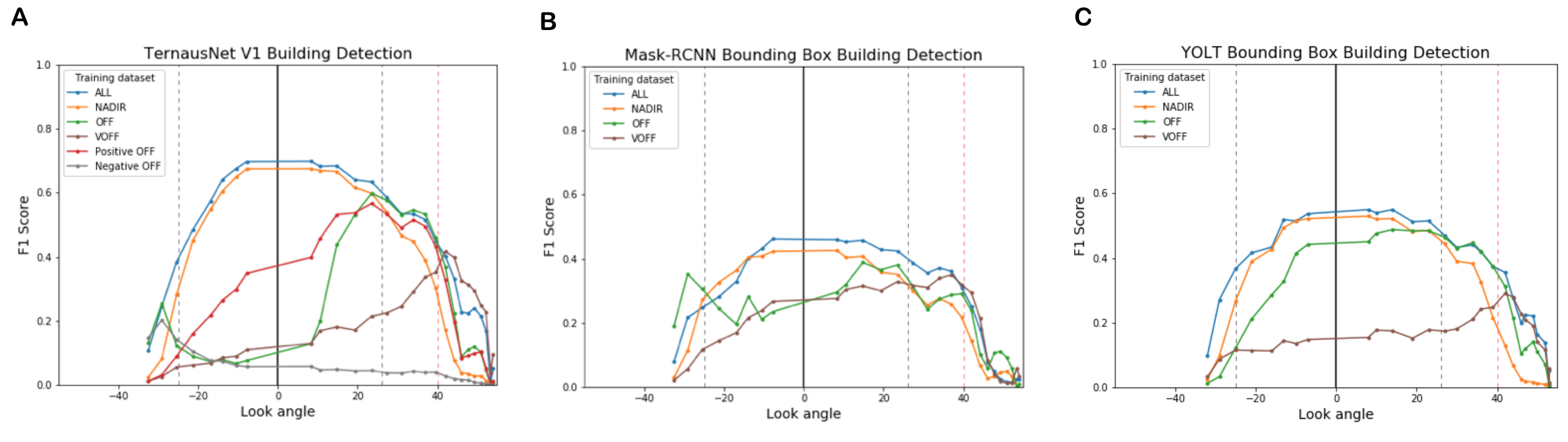}
    \vspace{-15pt}
    \caption{\textbf{Performance by look angle for various training subsets.} TernausNet (left), Mask R-CNN (middle), and YOLT (right) models, trained on ALL, NADIR, OFF, or VOFF, were evaluated in the building detection task and $F_1$ scores are displayed for each evaluation look angle. Imagery acquired facing South is represented as a negative number, whereas looks facing North are represented by a positive angle value. Additionally, TernausNet models trained only on North-facing OFF imagery (positive OFF) and South-facing OFF imagery (negative OFF) were evaluated on each angle to explore the importance of look direction.}
    \label{fig:gen_angles}
    \vspace{-10pt}
\end{figure*}

The state-of-the-art segmentation and object detection models we measured were challenged by this task. As shown in Table \ref{trainall}, TernausNet trained on all angles achieves $F_1 = 0.62$ on the nadir angles, which is on par with previous building segmentation results and competitions \cite{Etten2018SpaceNetAR, DeepGlobe18}. However, performance drops significantly for off-nadir ($F_1=0.43$) and very off-nadir ($F_1=0.22$) images. Other models display a similar degradation in performance. Example results are shown in Figure \ref{fig:comic_strip}.

\textbf{Directional asymmetry.} Figure \ref{fig:gen_angles} illustrates performance per angle for both segmentation and object detection models. Note that models trained on positive (north-facing) angles, such as Positive OFF (Red), fair particularly poorly when tested on negative (south-facing) angles. This may be due to the smaller dataset size, but we hypothesize that the very different lighting conditions and shadows make some directions intrinsically more difficult (Figure \ref{fig:challenges}C-D). This observation reinforces that developing models and datasets that can handle the diversity of conditions seen in overhead imagery in the wild remains an important challenge.

\textbf{Model architectures.} Interestingly, segmentation models designed specifically for overhead imagery (TernausNet and YOLT) significantly outperform general-purpose segmentation models for computer vision (U-Net, Mask R-CNN). These experiments demonstrate the value of specializing computer vision models to the target domain of overhead imagery, which has different visual, object density, size, and orientation characteristics.

\textbf{Effects of resolution.} OFF and VOFF images have lower base resolutions, potentially confounding analyses of effects due exclusively to look angle. To test whether resolution might explain the observed performance drop, we ran a control study with normalized resolution. We trained TernausNet on images from all look angles artificially reduced to the same resolution of $1.67m$, the lowest base resolution from the dataset. This model showed negligible change in performance versus the model trained on original resolution data (original resolution: $F_1 = 0.43$, resolution equalized: $F_1 = 0.41$) (Table \ref{tnet_res}). This experiment indicates that viewing angle-specific effects, not resolution, drive the decline in segmentation performance as viewing angle changes.

\begin{table}[bt]
\begin{center}
\begin{tabular}{lccc}
                      & \multicolumn{2}{c}{\textbf{Training Resolution}} \\
                      & Original      & {Equalized}   \\
                    \textbf{Test Angles} & (0.46-1.67 m) & 1.67 m \\
\hline
\hline
             NADIR   & 0.62          & 0.59  \\
             OFF     & 0.43          & 0.41  \\
             VOFF    & 0.22          & 0.22  \\
\hline
             Summary & 0.43          & 0.41  \\
\end{tabular}
\end{center}
\vspace{-10pt}
\caption{\textbf{TernausNet model trained on different resolution imagery.} Building footprint extraction performance for a TernausNet model trained on ALL original-resolution imagery (0.46 m ground sample distance (GSD) for $7.8^\circ$ to 1.67 m GSD at $54^\circ$), left, compared to the same model trained and tested on ALL imagery where every view is down-sampled to 1.67 m GSD (right). Rows display performance ($F_1$ score) on different angle bins. The original resolution imagery represents the same data as in Table \ref{trainall}. Training set imagery resolution had only negligible impact on model performance.}
\label{tnet_res}
\vspace{-10pt}
\end{table}


\textbf{Generalization to unseen angles.} Beyond exploring performance of models trained with many views, we also explored how effectively models could identify building footprints on look angles absent during training. We found that the TernausNet model trained only on NADIR performed worse on evaluation images from OFF (0.32) than models trained directly on OFF (0.44), as shown in Table \ref{tnet}. Similar trends are observed for object detection (Figure \ref{fig:gen_angles}). To measure performance on unseen angles, we introduce a generalization score $G$, which measures the performance of a model trained on $X$ and tested on $Y$, normalized by the performance of a model trained on $Y$ and tested on $Y$:
\begin{equation}
 G_Y = \frac{1}{N} \sum_{X} \, \frac{F_1(\text{train}=X,\text{test}=Y)}{F_1(\text{train}=Y,\text{test}=Y)}
 \label{eq:gen}
\end{equation}

\noindent This metric measures relative performance across viewing angles, normalized by the task difficulty of the test set. We measured $G$ for all our model/dataset combinations, as reported in Table \ref{generalization}. Even though the Mask R-CNN model has worse overall performance, the model achieved a higher generalization score ($G=0.78$) compared to TernausNet ($G=0.42$) as its performance did not decline as rapidly when look angle increased. Overall however, generalization scores to unseen angles were low, highlighting the importance of future study in this challenging task.

\begin{table}[tb]
\begin{center}
\begin{tabular}{lrrrr}
        & \multicolumn{4}{c}{\textbf{Training Angles}} \\
        \textbf{Test Angles} & All     & NADIR    & OFF    & VOFF   \\
\hline
\hline
NADIR   & 0.62    & 0.59     & 0.23   & 0.13   \\
OFF     & 0.43    & 0.32     & 0.44   & 0.23   \\
VOFF    & 0.22    & 0.04     & 0.13   & 0.27   \\
\hline
Summary & 0.43    & 0.32     & 0.26   & 0.21   \\
\end{tabular}
\end{center}
\vspace{-10pt}
\caption{\textbf{TernausNet model tested on unseen angles.} Performance ($F_1$ score) of the TernausNet model when trained on one angle bin (columns), and then tested on each of the three bins (rows). The model trained on NADIR performs worse on unseen OFF and VOFF views compared to models trained directly on imagery from those views.}
\label{tnet}
\vspace{-15pt}
\end{table}

\subsection{Effects of geography}
We broke down geographic tiles into Industrial, Sparse Residential, Dense Residential, and Urban bins, and examined how look angle influenced performance in each. We observed greater effects on residential areas than other types (Table S3). Testing models trained on MVOI with unseen cities\cite{Etten2018SpaceNetAR} showed almost no generalization (Table S4). Additional datasets with more diverse geographies are needed.

\section{Conclusion}
We present a new dataset that is critical for extending object detection to real-world applications, but also presents challenges to existing computer vision algorithms. Our benchmark found that segmenting building footprints from very off-nadir views was exceedingly difficult, even for state-of-the-art segmentation and object detection models tuned specifically for overhead imagery (Table \ref{trainall}). The relatively low $F_1$ scores for these tasks (maximum VOFF $F_1$ score of 0.22) emphasize the amount of improvement that further research could enable in this realm.

Furthermore, on all benchmark tasks we concluded that model generalization to unseen views represents a significant challenge. We quantify the performance degradation from nadir ($F_1$ = 0.62) to very off-nadir ($F_1$ = 0.22), and note an asymmetry between performance on well-lit north-facing imagery and south-facing imagery cloaked in shadows (Figure \ref{fig:challenges}C-D and Figure \ref{fig:gen_angles}). We speculate that distortions in objects, occlusion, and variable lighting in off-nadir imagery (Figure \ref{fig:challenges}), as well as the small size of buildings in general (Figure \ref{stats}), pose an unusual challenge for segmentation and object detection of overhead imagery.

The off-nadir imagery has a lower resolution than nadir imagery (due to simple geometry), which theoretically complicates building extraction for high off-nadir angles. However, by experimenting with imagery degraded to the same low $1.67m$ resolution, we show that resolution has an insignificant impact on performance (Table \ref{tnet_res}). Rather, variations in illumination and viewing angle are the dominant factors. This runs contrary to recent observations \cite{concurrent_SR}, which found that object detection models identify small cars and other vehicles better in super-resolved imagery.

\begin{table}[tb]
\begin{centering}
\setlength{\tabcolsep}{0.25em}
\begin{tabular}{llrrr}

             &          & \multicolumn{3}{c}{\textbf{Generalization Score} $\boldsymbol{G}$} \\
             \textbf{Task}    &     \textbf{Model}    & NADIR   & OFF    & VOFF  \\
\hline
\hline
Segmentation     & TernausNet & 0.45 & 0.43 & 0.37 \\
Segmentation     & U-Net         &  0.64    &  0.40    & 0.37   \\
Segmentation     & Mask R-CNN     & 0.60    & 0.90   & 0.84  \\
\hline
Detection & Mask R-CNN     & 0.64    & 0.92   & 0.76 \\
Detection & YOLT          & 0.57   & 0.68   & 0.44  \\
\end{tabular}
\end{centering}
\caption{\textbf{Generalization scores.} To measure segmentation model performance on unseen views, we compute a generalization score $G$ (Equation \ref{eq:gen}), which quantifies performance on unseen views normalized by task difficulty. Each column corresponds to a model trained on one angle bin.}
\label{generalization}
\vspace{-10pt}
\end{table}

The generalization score $G$ is low for the highest-performing, overhead imagery-specific models in these tasks (Table \ref{generalization}), suggesting that these models may be over-fitting to view-specific properties. This challenge is not specific to overhead imagery: for example, accounting for distortion of objects due to imagery perspective is an essential component of 3-dimensional scene modeling, or rotation prediction tasks \cite{lotter}. Taken together, this dataset and the $G$ metric provide an exciting opportunity for future research on algorithmic generalization to unseen views.

Our aim for future work is to expose problems of interest to the larger computer vision community with the help of overhead imagery datasets. While only one specific application, advances in enabling analysis of overhead imagery in the wild can concurrently solve broader tasks. For example, we had anecdotally observed that image translation and domain transfer models failed to convert off-nadir images to nadir images, potentially due to the spatial shifts in the image. Exploring these tasks as well as other novel research avenues will enable advancement of a variety of current computer vision challenges.

{\small
\bibliographystyle{ieee_fullname}
\bibliography{egbib}
}

\include{ICCV_supplement}

\end{document}

%% file: ICCV_supplement.tex

\begin{multicols}{2}[
\begin{center}
\huge{SpaceNet MVOI: a Multi-View Overhead Imagery Dataset \\ Supplementary Material}
\end{center}]
\end{multicols}
\appendix

\section{Dataset}

\subsection{Imagery details}

The images from our dataset were obtained from DigitalGlobe, with 27 different viewing angles collected over the same geographical region of Atlanta, GA. Each viewing angle is characterized as both an off-nadir angle and a target azimuth. We binned each angle into one of three categories (Nadir, Off-Nadir, and Very Off-Nadir) based on the angle (see Table \ref{dg}). Collects were also separated into South- or North-facing based on the target azimuth angle.

The imagery dataset comprises Panchromatic, Multi-Spectral, and Pan-Sharpened Red-Green-Blue-near IR (RGB-NIR) images The ground resolution of image varied depending on the viewing angle and the type of image (Panchromatic, Multi-spectral, Pan-sharpened). See Table \ref{res} for more details. All experiments in this study were performed using the Pan-Sharpened RGB-NIR image (with the NIR band removed, except for the U-Net model).

The imagery was uploaded into the spacenet-dataset AWS S3 bucket, which is publicly readable with no cost to download. Download instructions can be found at www.spacenet.ai/off-nadir-building-detection/.

\begin{table}[b]
\begin{center}
    \begin{tabular}{lrr}
        \hline
        \textbf{Image}  & \textbf{Resolution at 7.8$^\circ$} & \textbf{Resolution at 54$^\circ$} \\
        \hline
        \textbf{Panchromatic} & 0.46m/px & 1.67m/px\\
        \textbf{Multi-spectral} & 1.8m/px & 7.0m/px \\
        \textbf{Pan-sharpened} & 0.46m/px & 1.67m/px \\
        \hline
    \end{tabular}
    \caption{Resolution across different image types for two nadir angles.}
    \label{res}
\end{center}

\end{table}

\begin{table*}[tb]
\begin{tabular}{lrrrll}
\hline
\textbf{Catalog ID} & \textbf{Pan-sharpened Resolution} & \textbf{Look Angle} & \textbf{Target Azimuth Angle} & \textbf{Angle Bin} & \textbf{Look Direction} \\
\hline
 1030010003D22F00 & 0.48 & 7.8 & 118.4 & Nadir & South \\
 10300100023BC100 & 0.49 & 8.3 & 78.4 & Nadir & North \\
1030010003993E00 & 0.49 & 10.5 & 148.6 & Nadir & South \\
 1030010003CAF100 & 0.48 & 10.6 & 57.6 & Nadir & North \\
 1030010002B7D800 & 0.49 & 13.9 & 162 & Nadir & South \\
 10300100039AB000 & 0.49 & 14.8 & 43 & Nadir & North \\
 1030010002649200 & 0.52 & 16.9 & 168.7 & Nadir & South \\
 1030010003C92000 & 0.52 & 19.3 & 35.1 & Nadir & North \\
 1030010003127500 & 0.54 & 21.3 & 174.7 & Nadir & South \\
 103001000352C200 & 0.54 & 23.5 & 30.7 & Nadir & North \\
103001000307D800 & 0.57 & 25.4 & 178.4 & Nadir & South \\
 1030010003472200 & 0.58 & 27.4 & 27.7 & Off-Nadir & North \\
 1030010003315300 & 0.61 & 29.1 & 181 & Off-Nadir & South \\
 10300100036D5200 & 0.62 & 31 & 25.5 & Off-Nadir & North \\
 103001000392F600 & 0.65 & 32.5 & 182.8 & Off-Nadir & South \\
 1030010003697400 & 0.68 & 34 & 23.8 & Off-Nadir & North \\
 1030010003895500 & 0.74 & 37 & 22.6 & Off-Nadir & North \\
 1030010003832800 & 0.8 & 39.6 & 21.5 & Off-Nadir & North \\
 10300100035D1B00 & 0.87 & 42 & 20.7 & Very Off-Nadir & North \\
 1030010003CCD700 & 0.95 & 44.2 & 20 & Very Off-Nadir & North \\
1030010003713C00 & 1.03 & 46.1 & 19.5 & Very Off-Nadir & North \\
 10300100033C5200 & 1.13 & 47.8 & 19 & Very Off-Nadir & North \\
 1030010003492700 & 1.23 & 49.3 & 18.5 & Very Off-Nadir & North \\
 10300100039E6200 & 1.36 & 50.9 & 18 & Very Off-Nadir & North \\
 1030010003BDDC00 & 1.48 & 52.2 & 17.7 & Very Off-Nadir & North \\
1030010003193D00 & 1.63 & 53.4 & 17.4 & Very Off-Nadir & North \\
 1030010003CD4300 & 1.67 & 54 & 17.4 & Very Off-Nadir & North \\
\hline
\end{tabular}
\caption{DigitalGlobe Catalog IDs and the resolution of each image based upon off-nadir angle and target azimuth angle.}
\label{dg}
\end{table*}

\subsection{Dataset breakdown}
The imagery described above was split into three folds: 50\% in a training set, 25\% in a validation set, and 25\% in a final test set. $900 \times 900$-pixel geographic tiles were randomly placed in one of the three categories, with all of the look angles for a given geography assigned to the same subset to avoid geographic leakage. The full training set and building footprint labels as well as the validation set imagery were open sourced, and the validation set labels and final test imagery and labels were withheld as scoring sets for public coding challenges.

\section{Model Training}

\subsection{TernausNet}
The TernausNet model was trained without pre-trained weights roughly as described previously \cite{ternausnet}, with modifications. Firstly, only the Pan-sharpened RGB channels were used for training, and were re-scaled to 8-bit. $90^\circ$ rotations, X and Y flips, imagery zooming of up to 25\%, and linear brightness adjustments of up to 50\% were applied randomly to training images. After augmentations, a $512 \times 512$ crop was randomly selected from within each $900 \times 900$ training chip, with one crop used per chip per training epoch. Secondly, as described in the Models section of the main text, a combination loss function was used with a weight parameter $\alpha=0.8$. Secondly, a variant of Adam incorporating Nesterov momentum \cite{nadam} with default parameters was used as the optimizer. The model was trained for 25-40 epochs, and learning rate was decreased 5-fold when validation loss failed to improve for 5 epochs. Model training was halted when validation loss failed to improve for 10 epochs.
\begin{table}[b]
\vspace{-20pt}
\begin{center}
\begin{tabular}{lrrr}
    \hline
    \textbf{Type} & \textbf{NADIR} & \textbf{OFF - NADIR} & \textbf{VOFF - NADIR} \\
    \hline
    \textbf{Industrial}
    &  $0.51$ &  $-0.13$ & $-0.28$  \\
    \textbf{Sparse Res}
    &  $0.57$ & $-0.19$ & $-0.37$  \\
    \textbf{Dense Res}
    &  $0.66$ & $-0.21$ & $-0.41$ \\
    \textbf{Urban}
    &  $0.64$ & $-0.13$ & $-0.30$  \\
    \hline
\end{tabular}
\end{center}
\caption{$F_1$ score for the model trained on all angles and evaluated evaluated on the nadir bins (NADIR), then the relative decrease in $F_1$ for the off-nadir and very off-nadir bins.}
\label{mvoi_geogs}
\end{table}
\subsection{U-Net}
The original U-Net \cite{Ronneberger:2015gk} architecture was trained for 30 epochs with Pan-Sharpened RGB+NIR 16-bit imagery, on a binary segmentation mask with a combination loss as described in the main text with $\alpha=0.5$. Dropout and batch normalization were used at each layer, with dropout with $p=0.33$. The same augmentation pipeline was used as with TernausNet. An Adam Optimizer \cite{adam} was used with learning rate of 0.0001 was used for training.

\subsection{YOLT}
The You Only Look Twice (YOLT) model was trained as described previously \cite{yolt}. Bounding box training targets were generated by converting polygon building footprints into the minimal un-oriented bounding box that enclosed each polygon.

\subsection{Mask R-CNN}
The Mask R-CNN model with the ResNet50-C4 backbone was trained as described previously \cite{He_2017} using the same augmentation pipeline as TernausNet. Bounding boxes were created as described above for YOLT.

\section{Geography-specific performance}
\subsection{Distinct geographies within SpaceNet MVOI}
We asked how well the TernausNet model trained on SpaceNet MVOI performed both within and outside of the dataset. First, we broke down the test dataset into the four bins represented in main text Figure 1: Industrial, Sparse Residential, Dense Residential, and Urban, and scored models within those bins (Table \ref{mvoi_geogs}).  We observed slightly worse performance in Industrials areas than elsewhere at nadir, but markedly stronger drops in performance in residential areas as look angle increased.

\subsection{Generalization to unseen geographies}
We also explored how models trained on SpaceNet MVOI performed on building footprint extraction from imagery from other geographies, in this case, the Las Vegas imagery from SpaceNet \cite{Etten2018SpaceNetAR}. After normalizing the Las Vegas (LV) imagery for consistent pixel intensities and channel order with SpaceNet MVOI, we predicted building footprints in LV imagery and scored prediction quality as described in Metrics. We also re-trained TernausNet on the LV imagery and examined building footprint extraction quality on the SpaceNet MVOI test set. Strikingly, neither model was able to identify building footprints in the unseen geographies, highlighting that adding novel looks angles does not necessarily enable generalization to new geographic areas.

\begin{table}[hbt]
\vspace{-5pt}
\begin{center}
\begin{tabular}{lccc}
    & & \multicolumn{2}{c}{\textbf{Test Set}} \\
    \cline{3-4}
    & & MVOI 7$^\circ$      & SN LV  \\
    \hline
    \multirow{2}{*}{\textbf{Training Set}}
    & MVOI ALL   & 0.68     & 0.01  \\
    & SN LV      & 0.00     & 0.62  \\
    \hline
\end{tabular}
\end{center}
\vspace{-10pt}
\caption{\textbf{Cross-dataset $F_1$.} Models trained on MVOI or SpaceNet Las Vegas \cite{Etten2018SpaceNetAR} were inferenced on held out imagery from one of those two geographies, and building footprint quality was assessed as described in Metrics.}
\label{gen}
\vspace{-5pt}
\end{table}

{\small
\bibliographystyle{ieee_fullname}
\bibliography{ICCV_supplement}
}